\documentclass[10pt,twocolumn,letterpaper]{article}

\usepackage[pagenumbers]{cvpr} 

\usepackage[dvipsnames]{xcolor}

\usepackage{times}
\usepackage[utf8]{inputenc} 
\usepackage[T1]{fontenc}    
\usepackage{url}            
\usepackage{booktabs}       
\usepackage{amsfonts}       
\usepackage{nicefrac}       
\usepackage{microtype}      
\usepackage{xcolor}         

\usepackage{graphicx}
\usepackage{amsmath,amssymb}
\usepackage{verbatim}
\usepackage{multirow}
\usepackage{marvosym}
\usepackage{makecell}
\usepackage{bm}
\usepackage{wrapfig}
\usepackage{caption}
\usepackage{twemojis}
\usepackage{utfsym}

\usepackage[nomargin,inline,draft]{fixme} 
\FXRegisterAuthor{yf}{ayf}{Yifan}

\usepackage[pagebackref=true,breaklinks=true,letterpaper=true,colorlinks,bookmarks=false]{hyperref}
\definecolor{cvprblue}{rgb}{0.21,0.49,0.74}

\def\confName{CVPR}
\def\confYear{2024}

\title{Editable Scene Simulation for Autonomous Driving via Collaborative LLM-Agents}

\author{%
  Yuxi Wei$^1\footnotemark[1]$ \quad Zi Wang$^3\footnotemark[1]$ \quad Yifan Lu$^1\footnotemark[1]$ \quad Chenxin Xu$^1\footnotemark[1]$ \\ Changxing Liu$^1$ \quad Hao Zhao$^4$ \quad Siheng Chen$^{1,2}$ \quad Yanfeng Wang$^{1,2}$\\~
\small $^1$ Shanghai Jiao Tong University \quad
$^2$ Shanghai AI Laboratory \\
\small $^3$ Carnegie Mellon University \quad $^4$ Tsinghua University\\
 \texttt{\footnotesize \{wyx3590236732, yifan\_lu, xcxwakaka, cx-liu\}@sjtu.edu.cn}\\
 \texttt{\footnotesize \{sihengc,  wangyanfeng\}@sjtu.edu.cn}\\
 \texttt{\footnotesize ziwang2@andrew.cmu.edu}
 \texttt{\footnotesize zhaohao@air.tsinghua.edu.cn}
}

\begin{document}
\maketitle
\renewcommand{\thefootnote}{\fnsymbol{footnote}}
\footnotetext[1]{Equal contribution.}

\begin{abstract}
Scene simulation in autonomous driving has gained significant attention because of its huge potential for generating customized data. However, existing editable scene simulation approaches face limitations in terms of user interaction efficiency, multi-camera photo-realistic rendering and external digital assets integration. 
To address these challenges, this paper introduces~\textit{ChatSim}, the first system that enables editable photo-realistic 3D driving scene simulations via natural language commands with external digital assets. To enable editing with high command flexibility,~\textit{ChatSim} leverages a large language model (LLM) agent collaboration framework. To generate photo-realistic outcomes,~\textit{ChatSim} employs a novel multi-camera neural radiance field method. Furthermore, to unleash the potential of extensive high-quality digital assets,~\textit{ChatSim} employs a novel multi-camera lighting estimation method to achieve scene-consistent assets' rendering. Our experiments on Waymo Open Dataset demonstrate that~\textit{ChatSim} can handle complex language commands and generate corresponding photo-realistic scene videos. Code can be accessed at: \href{https://github.com/yifanlu0227/ChatSim}{https://github.com/yifanlu0227/ChatSim}.



\end{abstract}    
\section{Introduction}
\label{sec:intro}


 \quad Perception~\cite{chen2015deepdriving,chen2017multi,wu2017squeezedet,chen2016monocular,caesar2020nuscenes,li2020rtm3d} is the window of an autonomous vehicle into the external environment. To ensure the robustness of the vehicle's perceptual capabilities during both training and testing phases, it necessitates the collection of high-quality perception data in substantial volumes \cite{chang2019argoverse,sun2020scalability}. However, the operation of a fleet for the acquisition of real-world data often incurs prohibitive expenses, particularly for specialized or customized requirements. For instance, in the aftermath of an accident or intervention involving an autonomous vehicle, it is imperative to test the vehicle's perception system across a spectrum of similar scenarios. While replicating such scenario data from real-world instances is nearly impossible due to the uncontrollability of actual scenes \cite{rong2020lgsvl,bergamini2021simnet}, customized scene simulation emerges as a vital and feasible alternative. It enables the precise modeling of specific conditions without high costs and logistical complexities of real-world data collection \cite{amini2020learning,yang2020surfelgan}.

\begin{figure}[t] 
\centering 
\includegraphics[width=0.47\textwidth]{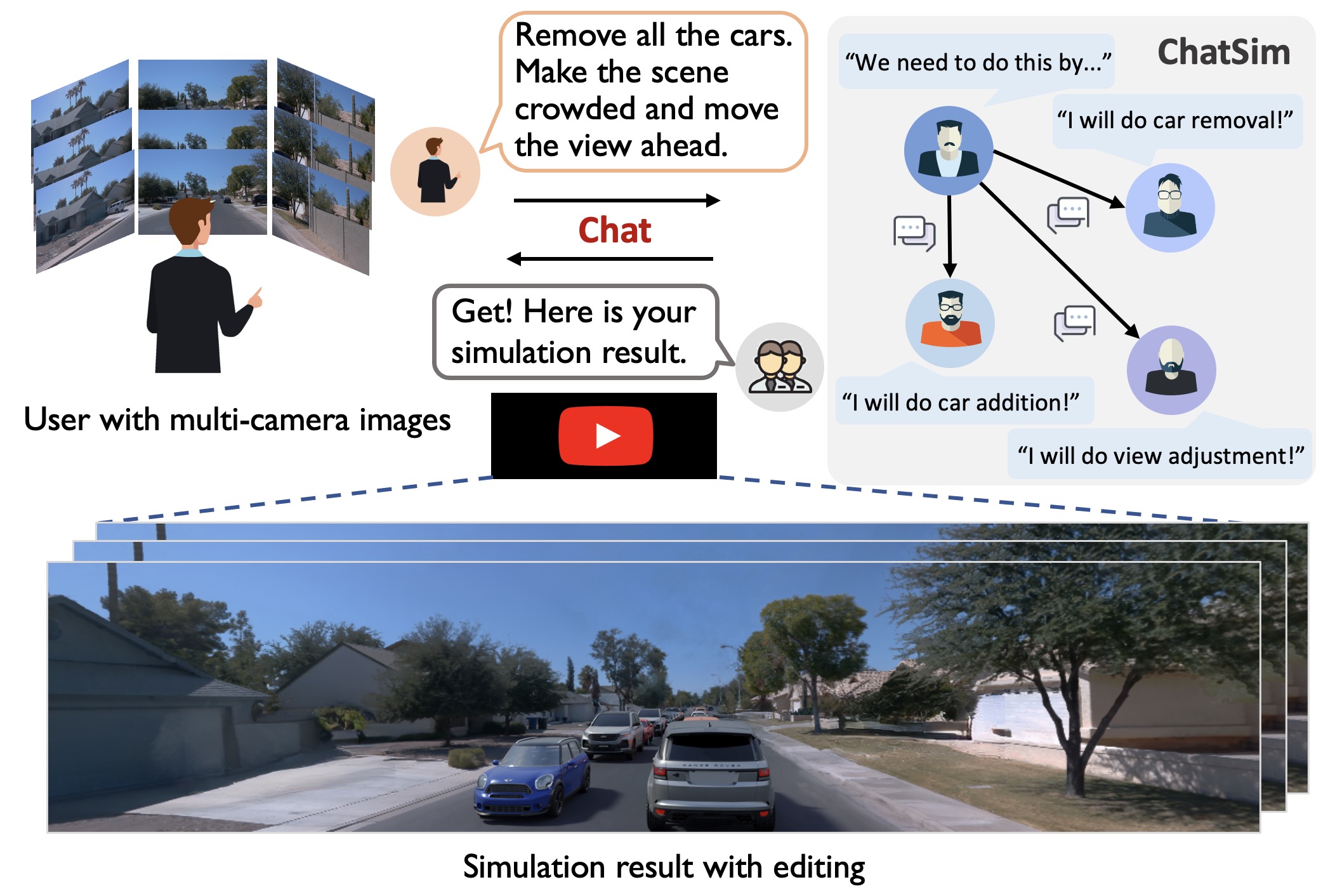} 
\vspace{-3.5mm}
\caption{\small \textit{ChatSim} enables the editing of photo-realistic 3D driving scene simulations via language commands. 
}
\label{Fig:teaser} 
\vspace{-6mm}
\end{figure}

To effectively simulate customized driving scenes, we identify three key properties as fundamental. First, the simulation should be capable of following sophisticated or abstract demands, thereby 
facilitating the production. Second, the simulation should generate photo-realistic, view-consistent outcomes, which allow for the closest approximation to vehicle observations in real-world scenarios. Third, it should allow for the integration of external digital assets \cite{miric2019protecting,banta2016property} with their photo-realistic textures and materials while fitting the lighting conditions. This capability would unlock the potential for data expansion by incorporating a wide array of external digital assets, satisfying customized needs.

A vast array of significant works have been proposed for scene simulation, yet they fail to meet all three of these requirements. Traditional graphics engines, such as CARLA~\cite{dosovitskiy2017carla} and UE \cite{UE}, offer editable virtual environments with external digital assets, but the data realism is restricted by asset modeling and rendering qualities. Image generation based methods, such as BEVControl \cite{yang2023bevcontrol}, DriveDreamer \cite{wang2023drivedreamer}, MagicDrive \cite{gao2023magicdrive}, can generate realistic scene images based on various control signals, including BEV maps, bounding boxes and camera poses. However, they struggle to maintain view consistency and face challenges in importing external digital assets due to the absence of 3D spatial modeling. Rendering-based methods have been proposed to obtain photo-realistic and view-consistent scene simulation. Notable examples like UniSim \cite{yang2023unisim} and MARS \cite{wu2023mars} come equipped with a suite of scene-editing tools. However, these systems require extensive user involvement in every trivial editing step via code implementation, which is ineffective when performing the editing. Furthermore, while they handle vehicles in observed scenarios effectively, their inability to support external digital assets restricts opportunities for data expansion and customization.

To fulfill the identified requirements, we introduce~\textit{ChatSim}, the first system that enables editable photo-realistic 3D driving scene simulations via natural language commands with external digital assets. To use~\textit{ChatSim}, users simply engage in a conversation with the system, issuing commands through natural language without any involvement in intermediate simulation steps; see Figure \ref{Fig:teaser} for illustration.

To address complex or abstract user commands effectively,~\textit{ChatSim} adopts a large language model (LLM)-based multi-agent collaboration framework. The key idea is to exploit multiple LLM agents, each with a specialized role, to decouple an overall simulation demand into specific editing tasks, thereby mirroring the task division and execution typically founded in the workflow of a human-operated company. This workflow offers two key advantages for scene simulation. First, LLM agents' ability to process human language commands allows for intuitive and dynamic editing of complex driving scenes, enabling precise adjustments and feedback. Second, the collaboration framework enhances simulation efficiency and accuracy by distributing specific editing tasks among specialized agents, ensuring detailed and realistic simulations with improved task completion rates.

To generate photo-realistic outcomes, we propose McNeRF in ~\textit{ChatSim}, a novel neural radiance field method that incorporates multi-camera inputs, offering a broader scene rendering. This integration fully exploits camera setups on vehicles but raises two significant challenges: camera pose misalignment due to asynchronized trigger times and brightness inconsistency due to different camera exposure times. To address camera pose misalignment, McNeRF uses a multi-camera alignment to reduce extrinsic parameter noises, ensuring rendering quality. To address brightness inconsistency, McNeRF integrates the critical exposure times to recover scene radiance in HDR,  markedly mitigating the issue of color discrepancies at the intersections of two camera images with different exposure times.


To import external digital assets with their realistic textures and materials, we propose McLight, a novel multi-camera lighting estimation that blends skydome and surrounding lighting. Our skydome estimation restores accurate sun behavior with peak intensity residual connection, enabling the rendering of prominent shadows. For surrounding lighting, McLight queries McNeRF to achieve complex location-specific illumination effects, like those in the tree shade with sunlight being blocked. This significantly improves the rendering realism of the integrated 3D assets.

 
We conduct extensive experiments on the Waymo autonomous driving dataset and show that ~\textit{ChatSim} generates photo-realistic customized perception data including dangerous corner cases according to various human language commands. Our method is compatible with mixed, highly-abstract and multi-round commands. Our method achieves SoTA performance with an improvement of 4.5\% in photo-realism with a wide-angle rendering. Moreover, we demonstrate our lighting estimation outperforms the SoTA methods both qualitatively and quantitatively, reducing the intensity error and angular error by 57.0\% and 9.9\%. 




\section{Related Work}
\vspace{-1mm}
\textbf{Scene simulation for autonomous driving.}
Current scene simulation methods can be generally divided into three categories: graphics engines, image generation, and scene rendering. Graphics engines, such as CARLA \cite{dosovitskiy2017carla}, AirSim \cite{shah2018airsim}, OpenScenario Editor \cite{openscanerio}, 51Sim-One \cite{51sim} and RoadRunner \cite{crescenzi2001roadrunner}, create a virtual world for simulating a wide range of driving scenarios. However, there exists a significant domain gap between the virtual world and reality. Image generation methods can generate realistic scene images based on different control signals, such as HD maps \cite{swerdlow2023street,gao2023magicdrive,li2023drivingdiffusion}, sketch layout~\cite{yang2023bevcontrol},  bounding boxes~\cite{li2023drivingdiffusion,wang2023drivedreamer,gao2023magicdrive}, text~\cite{li2023drivingdiffusion,wang2023drivedreamer,gao2023magicdrive,hu2023gaia} and driving actions~\cite{wang2023drivedreamer,hu2023gaia}. However, these approaches can hardly maintain scene consistency.  
To obtain a coherent driving scene, methods based on scene rendering target to reconstruct the 3D scene.  READ~\cite{li2023read} employs point clouds and uses a U-Net to render images. With the rapid development of Neural Radiance Field (NeRF)~\cite{mildenhall2021nerf, barron2021mip, barron2022mip, muller2022instant, sun2022direct, wang2023f2}, several works ~\cite{xie2023s, guo2023streetsurf, yang2023unisim, wu2023mars, ost2021neural, turki2023suds, kundu2022panoptic, ost2022neural} also exploit NeRFs
to model cars and static street backgrounds in outdoor environments. Moreover, notable examples like UniSim \cite{yang2023unisim} and MARS \cite{wu2023mars} come equipped with a suite of scene-editing tools.
However, these methods require extensive user involvement in intermediate editing steps and they fail to support external digital assets for data expansion. 
In this work, we propose \textit{ChatSim} that achieves automatic simulation editing via language commands and integrates external digital assets to enhance realism and flexibility. In \textit{ChatSim}, we integrate McNeRF, a novel neural radiance field designed to leverage multi-camera inputs for high-fidelity rendering.

\noindent\textbf{Lighting estimation.} Lighting estimation focuses on assessing the illumination conditions of a real-world environment to seamlessly integrate digital objects. Early methods \cite{lalonde2014lighting,lalonde2010sun} for outdoor environments use explicit cues like detected shadows on the ground. Recent works usually adopt learning-based approaches \cite{garon2019fast,hold2019deep,legendre2019deeplight,hold2017deep,li2018learning,zhang2019all} by predicting different lighting representations like spherical lobes \cite{boss2020two,li2018learning}, light probes \cite{legendre2019deeplight}, environment
map \cite{sengupta2019neural,somanath2021hdr}, HDR sky model \cite{hold2019deep,zhang2019all,wang2022neural} and lighting volume \cite{wang2022neural}. However, few of them consider multi-camera input, which is common for driving scenarios. In this paper, we propose a novel multi-camera lighting estimation method, McLight, combining with our McNeRF, to estimate a wider range of lighting and obtain the spatially-varying lighting effects of assets.

\begin{table}[t]
\center
\scriptsize
\setlength\tabcolsep{2pt}
\begin{tabular}{c|cccccccc}
\hline
Method     & \makecell{Photo-\\realistic}   & {Dim.} & \makecell{Multi-\\camera}                                                 & Editable & \makecell{External \\ assets} & Language & \makecell{Open- \\source}                             \\ \hline
CARLA~\cite{dosovitskiy2017carla} & \scalebox{0.75}{\usym{2613}}  &  3D& \checkmark 
& \checkmark & \checkmark & \scalebox{0.75}{\usym{2613}} & \checkmark \\
AirSim~\cite{shah2018airsim} & \scalebox{0.75}{\usym{2613}}  &  3D& \checkmark 
& \checkmark & \checkmark & \scalebox{0.75}{\usym{2613}}&  \checkmark \\
OpenScenario \cite{openscanerio} & \scalebox{0.75}{\usym{2613}}  &  3D& \checkmark 
& \checkmark  & \checkmark & \scalebox{0.75}{\usym{2613}}&  \checkmark \\
51Sim-One \cite{51sim}& \scalebox{0.75}{\usym{2613}}  &  3D& \checkmark 
& \checkmark & \checkmark & \scalebox{0.75}{\usym{2613}}& \scalebox{0.75}{\usym{2613}} \\
RoadRunner \cite{crescenzi2001roadrunner} & \scalebox{0.75}{\usym{2613}}  &  3D& \checkmark 
& \checkmark  & \checkmark& \scalebox{0.75}{\usym{2613}}& \checkmark \\
BEVGen~\cite{swerdlow2023street}     & \checkmark                & 2D                                      & \checkmark          & \checkmark      &  \scalebox{0.75}{\usym{2613}}       &     \scalebox{0.75}{\usym{2613}}                 &   \checkmark                            \\ 
BEVControl~\cite{yang2023bevcontrol}     & \checkmark            & 2D                          & \checkmark          & \checkmark     &  \scalebox{0.75}{\usym{2613}}      & \scalebox{0.75}{\usym{2613}}                       &      \scalebox{0.75}{\usym{2613}}                         \\ 
DriveDreamer~\cite{wang2023drivedreamer}      & \checkmark         & 2D                                   & \checkmark                     & \checkmark   &  \scalebox{0.75}{\usym{2613}}        &  \checkmark  &         \scalebox{0.75}{\usym{2613}}                            \\ 
DrivingDiffusion~\cite{li2023drivingdiffusion}    & \checkmark       & 2D                          & \checkmark                           & \checkmark   &      \scalebox{0.75}{\usym{2613}}   & \checkmark                          & \scalebox{0.75}{\usym{2613}}                             \\ 
GAIA-1~\cite{hu2023gaia}         & \checkmark            & 2D                  & \scalebox{0.75}{\usym{2613}}                  &    \checkmark    & \scalebox{0.75}{\usym{2613}}   & \checkmark                                & \scalebox{0.75}{\usym{2613}}                             \\ 
MagicDrive~\cite{gao2023magicdrive}     & \checkmark            & 2D                          & \checkmark  &\checkmark      & \scalebox{0.75}{\usym{2613}}     & \scalebox{0.75}{\usym{2613}}                               &   \scalebox{0.75}{\usym{2613}}                               \\ 
READ~\cite{li2023read}       & \checkmark                & 3D                          & \scalebox{0.75}{\usym{2613}}          & \scalebox{0.75}{\usym{2613}}         &\scalebox{0.75}{\usym{2613}}  &   \scalebox{0.75}{\usym{2613}}                                     & \checkmark                             \\ 
Neural SG~\cite{ost2021neural}    
& \checkmark    & 3D                          & \scalebox{0.75}{\usym{2613}}          & \checkmark        &\scalebox{0.75}{\usym{2613}}      & \scalebox{0.75}{\usym{2613}}                                     & \checkmark               \\ 
Neural PLF~\cite{ost2022neural}
& \checkmark  & 3D                          & \scalebox{0.75}{\usym{2613}}       & \scalebox{0.75}{\usym{2613}}          &\scalebox{0.75}{\usym{2613}}      & \scalebox{0.75}{\usym{2613}}                 & \checkmark                             \\ 
S-NeRF~\cite{xie2023s}       & \checkmark              & 3D                                 & \checkmark           & \scalebox{0.75}{\usym{2613}}    & \scalebox{0.75}{\usym{2613}}    & \scalebox{0.75}{\usym{2613}}                       & \checkmark                            \\ 
UniSim~\cite{yang2023unisim}      & \checkmark               & 3D                                 & \scalebox{0.75}{\usym{2613}}           & \checkmark    & \scalebox{0.75}{\usym{2613}}     & \scalebox{0.75}{\usym{2613}}                                    & \scalebox{0.75}{\usym{2613}}                             \\ 
MARS~\cite{wu2023mars}      & \checkmark                 & 3D                          & \scalebox{0.75}{\usym{2613}}            & \checkmark     &\scalebox{0.75}{\usym{2613}}        & \scalebox{0.75}{\usym{2613}}                 & \checkmark                              \\ 
\textbf{ChatSim (Ours)}       & \checkmark                & 3D                                   & \checkmark              & \checkmark     &\checkmark        & \checkmark                     & \checkmark                \\ \hline
\end{tabular}
\vspace{-3mm}
\caption{\small Comparison of existing and proposed methods for autonomous driving simulation.}
\label{tab:method_comparison}
\vspace{-4mm}
\end{table}



\noindent\textbf{Large language model and collaborative framework.}
Large Language Models (LLMs) are AI systems trained on extensive data to understand, generate, and respond to human language. GPT \cite{llm_gpt3} is a pioneering work to generate human-like content. The following updated versions GPT-3.5 \cite{chatgpt} and GPT-4 \cite{llm_gpt4}, provide more intelligent capabilities like chatting, browsing and coding. Notable other large language models include InstructGPT \cite{ouyang2022training}, LLaMA \cite{touvron2023llama} and PaLM \cite{chowdhery2022palm,anil2023palm}. Based on LLM, many works \cite{wang2023unleashing,du2023improving,zhuge2023mindstorms,hao2023chatllm,akata2023playing} improve the problem-solving abilities by integrating communication among multiple agents. \cite{hong2023metagpt} and \cite{wu2023autogen} define a group of well-organized agents to form operating procedures with conversation and code programming. 
In this paper, we exploit the power of collaborative LLM agents in simulation for 
autonomous driving, enabling the various editing of 3D scenes via language commands.

\section{Collaborative LLM-Agents for Editing}
\begin{figure*}[t] 
\centering 
\includegraphics[width=0.99\textwidth]{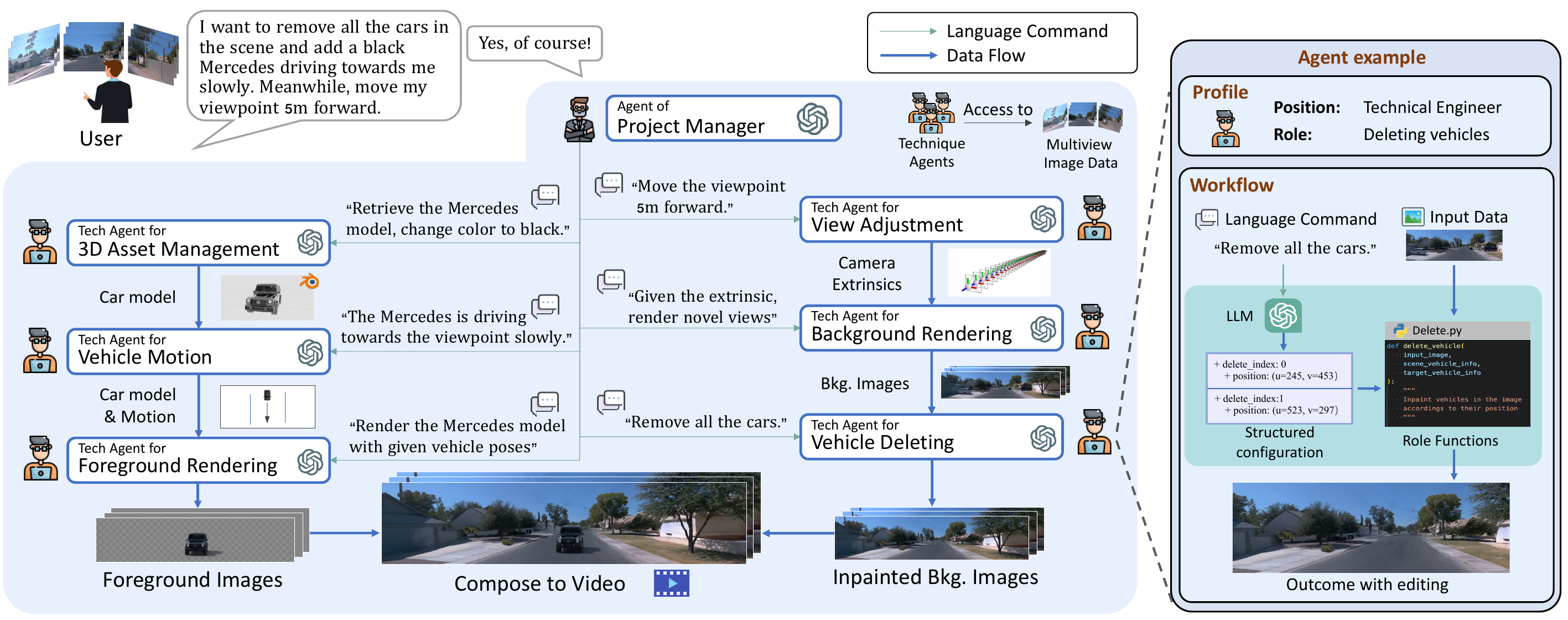} 
\vspace{-3mm}
\caption{\small \textbf{\textit{ChatSim} system overview.} The system exploit multiple collaborative LLM agents with specialized roles to decouple an overall demand into specific editing tasks. Each agent equips an LLM and corresponding role functions to interpret and execute its specific tasks.   }
\label{Fig:overview} 
\vspace{-4mm}
\end{figure*}

\vspace{-1mm}
\label{sec:3}
\quad
The \textit{ChatSim} system analyzes specific user commands and returns a video that meets customized needs; see Figure \ref{Fig:overview}. Since user commands could be abstract and sophisticated, it requires the system to have flexible task-handling ability. Directly applying a single LLM agent struggles with multi-step reasoning and cross-referencing. To address this, we design a series of collaborative LLM agents, where each agent is responsible for a unique aspect of the editing task. 

\begin{figure}[t] 
\centering 
\includegraphics[width=0.48\textwidth]{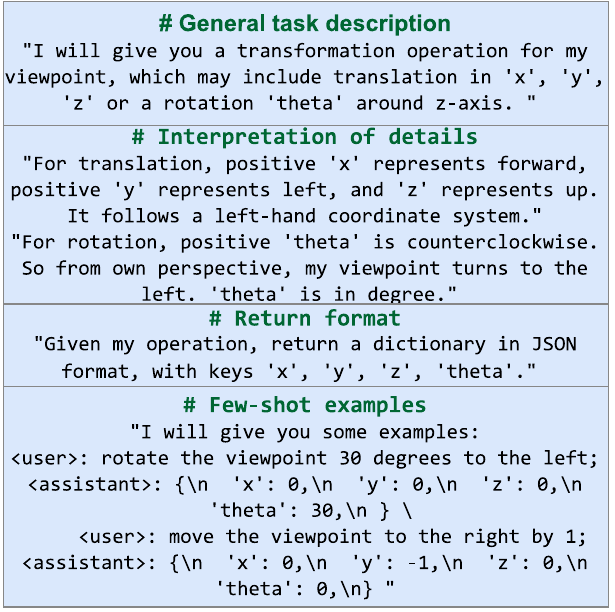} 
\vspace{-3mm}
\caption{\small  Prompt example of view adjustment agent.}
\label{Fig:prompt} 
\vspace{-4mm}
\end{figure}

\subsection{Specific Agent's Functionality}
\vspace{-1mm}
\quad
Agents in \textit{ChatSim} comprise two key components: a Large Language Model (LLM) and the corresponding role functions. The LLM is responsible for understanding the received commands while the role functions process the received data. Each agent is equipped with unique LLM prompts and role functions tailored to their specific duties within the system. To accomplish their tasks, agents first convert the received commands to a structured configuration using LLM with the assistance of prompts. 
Then the role functions utilize the structured configuration as parameters to process the received data and produce the desired outcomes; see an agent example on the right side of Figure \ref{Fig:overview}. This workflow endows agents with both language interpretation capabilities and precise execution capabilities.

\noindent\textbf{Project Manager Agent.} The project manager agent decomposes direct commands into clear natural language instructions dispatched to other editing agents.
To equip the project manager agent with the capability of command decomposition, we design a series of prompts for its LLM. The core idea of the prompts is to describe the action set, give the overall goal, and define the output form with examples; 
The role functions send the decomposed instructions to other agents for editing. The presence of the project manager agent enhances the system's robustness in interpreting various inputs and streamlines operations for clarity and fine granularity.

\noindent\textbf{Tech agent for view adjustment.}
The view adjustment agent generates suitable extrinsic camera parameters. The LLM in the agent translates the natural language instructions for viewpoint adjustment into movement parameters to the target viewpoint's position and angle. In role functions, 
these movement parameters are turned into transformation matrices required by the extrinsic, which are then multiplied by the original parameters to yield a new viewpoint. 

\noindent\textbf{Tech agent for background rendering.}
The background rendering agent renders the scene background based on multi-camera images. The LLM receives the rendering command and then operates the role functions for rendering. Notably, in role functions, we specifically integrate a novel neural radiance field method (McNeRF) taking multi-camera inputs and considering exposure time, solving the problem of blurring and brightness inconsistency in multi-camera rendering, see more details in Section \ref{subsec:bg_rendering}.



\noindent\textbf{Tech agent for vehicle deleting.}
The vehicle deleting agent removes specified vehicles from the background. It first identifies current vehicle attributes like 3D bounding boxes and colors from given scene information or results from a scene perception model like \cite{liu2023bevfusion}. The LLM gathers attributes of the vehicles and performs matching with user requests. Upon confirming the targeted vehicles, it employs a per-frame inpainting model as the role functions, such as latent diffusion methods~\cite{rombach2021highresolution}, to effectively delete them from the scene. 

\noindent\textbf{Tech agent for 3D asset management.}
The 3D asset management agent selects and modifies 3D digital assets according to user specifications. It constructs and maintains a 3D digital asset bank; see our bank details in the Appendix. To facilitate the addition of various objects, the agent first uses LLM to select the most suitable asset by key attributes matching with the requirements, such as color and type. If the matching is not perfect, the agent could modify the asset through its role functions like changing the color.


\noindent\textbf{Tech agent for vehicle motion.} 
The vehicle motion agent creates the initial places and subsequent motions of vehicles following the requests. Existing vehicle motion generation methods cannot directly generate motion purely from text and the scene map. To solve the problem, here we propose a novel text-to-motion method. The key idea is linking a placement and planning module as role functions with LLMs to extract and turn motion attributes into coordinates.
Motion attributes include position attributes (e.g., distance, direction) and movement attributes (e.g., speed, action). 
For the placement module, we endow each lane node in the lane map with its attributes to match with the position attributes. The planning module plans the vehicle's approximate destination lane node and then plans the intermediate trajectory by fitting the Bezier curves. 
We also add trajectory tracking~\cite{xu2023drl} to fit vehicle dynamics; see more details in the Appendix.

\noindent\textbf{Tech agent for foreground rendering.}
The foreground rendering agent integrates camera extrinsic infomation, 3D assets, and motion information to render foreground objects in the scene. Notably, to seamlessly integrate the external assets with the current scene, we design a multi-camera lighting estimation method (McLight) into the role functions, coupling with McNeRF. The estimated illumination is then utilized by Blender API to generate foreground images. The detailed technical aspects will be elaborated in Section \ref{sec:fg_rendering}.

\subsection{Agent Collaboration Workflow}
\vspace{-1mm}
\quad
Agents with tailored functions collaboratively work together to edit based on user commands. The project manager orchestrates and dispatches instructions to editing agents. The editing agents form two teams: background generation and foreground generation. For background generation, the background rendering agent generates rendered images using the extrinsic parameters from the view adjustment agent, followed by inpainting by the vehicle deleting agent. For foreground generation, the foreground rendering agent renders the images using the extrinsic parameters from the view adjustment agent, selected 3D assets from 3D asset management agent, and generated motions from vehicle motion agent. Finally, the foreground and background images are composed to create and deliver a video to the user. The editing information in each agent's configuration is recorded by the project manager agent for possible multi-round editings.

\section{Novel Rendering Methods}
\vspace{-1mm}
\quad
Based on the collaborative LLM agents framework introduced in Section~\ref{sec:3}, this section presents two novel rendering techniques to enhance photo-realism in simulations. To tackle the rendering challenges caused by multiple cameras, we propose multi-camera neural radiance field (McNeRF), a novel NeRF model considering the varied camera exposure times for visual consistency. To render realistic external digital assets with location-specific lighting and accurate shadows, we propose McLight, a novel hybrid lighting estimation method combined with our McNeRF. Note that McNeRF and McLight are leveraged by the background rendering agent and the foreground rendering agent, respectively.

\subsection{McNeRF for Background Rendering} 
\label{subsec:bg_rendering}
\quad
An autonomous vehicle typically equips multiple cameras to achieve a comprehensive perception view. However, this poses challenges for NeRF training due to the misaligned multi-camera poses from asynchronized camera trigger times and the brightness inconsistency originating from different exposure times. To address these challenges, the proposed McNeRF uses two techniques: multi-camera alignment and brightness-consistent rendering.

\noindent\textbf{Multi-camera alignment.} 
Autonomous vehicles, despite having a localization module for accurate camera poses, face challenges with asynchronous trigger times across multiple cameras. To align camera extrinsics for NeRF training, our core idea is to leverage a consistent spatial coordinate system provided by Agisoft Metashape~\cite{agisoft2019metashape} to align the images captured by multiple cameras at different timestamps. 


Specifically, let $\mathcal{I}^{(i,k)}$ and $\mathcal{\xi}^{(i,k)}$ be the image captured by the $i$th camera at the $k$th trigger and the corresponding camera pose in the vehicle's global coordinate space, respectively. We first input all images into Metashape for recalibration.
The aligned camera pose is then obtained as:
\begin{equation*}
\setlength\abovedisplayskip{2pt}
\setlength\belowdisplayskip{2pt}
    \widehat{\xi}^{(i,k)} = T_{M \rightarrow G} \cdot \xi_{M}^{(i,k)},
\end{equation*}
where $\xi_{M}^{(i,k)}$ denotes the recalibrated camera pose in the Metashape's unified spatial coordinate space, and $T_{M \rightarrow G}$ is the transformation from the Metashape's coordinate space to the vehicle's global coordinate space. After alignment, the pose noise can be significantly reduced. Then, the aligned camera pose $\widehat{\xi}^{(i,t)} $ can be used to generate the origins and directions of rays for McNeRF, enabling high-fidelity rendering. The aligned pose can also facilitate the foreground rendering agent's operations.



\begin{figure}[t] 
\centering
\includegraphics[width=0.47\textwidth]{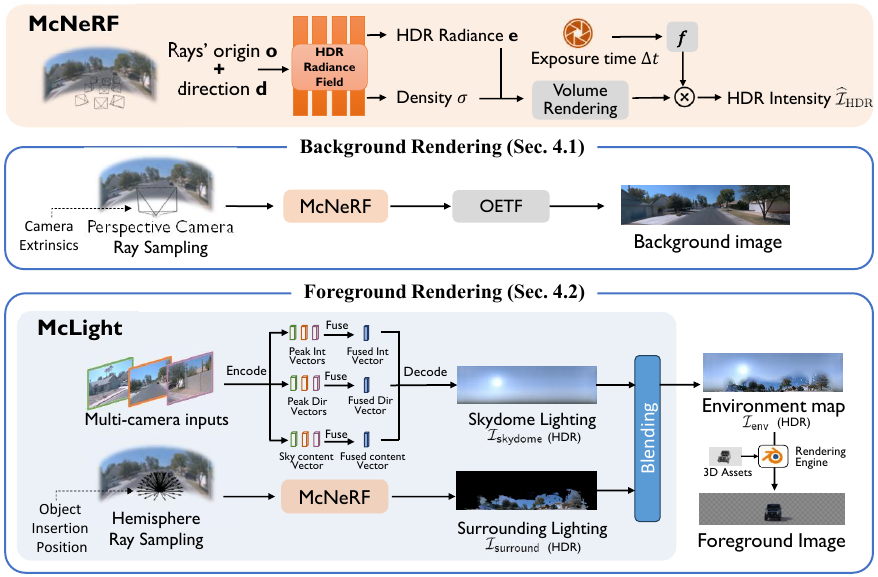}
\vspace{-3mm}
\caption{\small Rendering framework. The main components include McNeRF and McLight. Background rendering uses McNeRF to predict HDR pixel value and convert it to LDR with sRGB OETF. McLight includes a skydome lighting estimation network and adopts McNeRF to generate surrounding lighting.}
\label{fig:rendering_pipeline}
\vspace{-6mm}
\end{figure}

\noindent\textbf{Brightness-consistent rendering.} 
The exposure times of cameras can differ substantially, causing significant brightness differences across images, hindering the NeRF training. As shown in Figure~\ref{fig:rendering_pipeline}, McNeRF, addresses this by incorporating exposure times into HDR radiance fields, prompting brightness consistency.

We adopt F2-NeRF~\cite{wang2023f2} as our backbone model to handle the unbounded scene, sampling $K$ points along the ray $\mathbf{r}$ and estimating each point's HDR radiance $\mathbf{e}_k$ and density $\sigma_k$. The HDR light intensity is then calculated as:
\begin{equation}
\setlength\abovedisplayskip{2pt}
\setlength\belowdisplayskip{2pt}
    \widehat{\mathcal{I}}_{\rm HDR}(\mathbf{r}) = f(\Delta t) \cdot  \sum^K_{k=1} T_k \alpha_k \mathbf{e}_k,
\label{eq:nerf}
\end{equation}
where $\alpha_k = 1-\text{exp}(-\sigma_k \delta_i)$ is the opacity, $\delta_i$ is the point sampling interval, $T_k = \prod^{k-1}_{i=0} (1-\alpha_i)$ is the accumulated transmittance and $\Delta t$ is the exposure time. The normalization function $f(\Delta t) = 1 + \epsilon (\Delta t - \mu)/{\sigma}$ is designed to stabilize training, where $\epsilon$ is a hyperparameter for scaling, $\mu$ and $\sigma$ are the mean and standard deviation of the exposure times of all images, respectively.

By predicting scene radiance in HDR and multiplying it by the exposure time, we recover the light intensity received by the sensor and tackle the inconsistent color supervision at the intersections of two camera images with distinct exposure times.
Moreover, the HDR light intensity outputted by McNeRF can provide scene-level illumination for foreground object rendering, a topic further discussed in Section~\ref{sec:fg_rendering}. 


To train the rendering network, we enforce the consistency of radiance between the rendered image (prediction) and the captured image (ground-truth). Given the ground-truth image $\mathcal{I}$, the loss function is then:
\begin{equation*}
\setlength\abovedisplayskip{2pt}
\setlength\belowdisplayskip{2pt}
\mathcal{L} = \frac{1}{|R|} \sum_{\mathbf{r} \in R} \left(\mathrm{OETF}\left(\widehat{\mathcal{I}}_{\rm HDR}(\mathbf{r}) \right) - \mathcal{I}(\mathbf{r}) \right)^2,
\end{equation*}
where $R$ represents the ray set and $\mathrm{OETF}(\cdot)$ is the sRGB opto-electronic transfer function (gamma correction)~\cite{international1999iec} that converts HDR light intensity to LDR colors.



\subsection{McLight for Foreground Rendering} 
\label{sec:fg_rendering}

\quad
To enrich the scene's content with substantial digital 3D assets, we employ Blender~\cite{blender} foreground virtual objects' rendering. A seamless insertion critically depends on accurately estimating the scene's illumination conditions. Thus, as shown in Figure~\ref{fig:rendering_pipeline}, we propose McLight, a novel hybrid lighting estimation consisting of skydome lighting and surrounding lighting.

\noindent\textbf{Skydome lighting estimation.}
Estimating skydome lighting from images is challenging for restoring accurate sun behavior. To achieve this, we propose a novel residual connection from the estimated peak intensity to the HDR reconstruction to address over-smoothing output. Further, we adopt a self-attention mechanism to fuse multi-camera inputs, capturing complementary visual cues. 

Here we employ a two-stage process. In the first stage, we train an autoencoder to reconstruct the corresponding HDR panorama from an LDR panorama. Following ~\cite{wang2022neural}, the encoder transforms the LDR skydome panorama into three intermediate vectors, including the peak direction vector $\mathbf{f}_{\rm dir} \in \mathbb{R}^{3}$, the intensity vector $\mathbf{f}_{\rm int} \in \mathbb{R}^{3}_{+}$, and the sky content vector $\mathbf{f}_{\rm content} \in \mathbb{R}^{64}$. However, as HDR intensity behaves like an impulse response at its peak position, with pixel values thousands of times higher than its neighbors,  it is difficult for the decoder to recover such patterns. To tackle this, we design a residual connection that injects $\mathbf{f}_{\rm int}$ into the decoded HDR panorama with a spherical Gaussian lobe attenuation. This explicitly restores the peak intensity of the sun in the reconstructed HDR panorama, allowing us to render strong shadows for virtual objects. 


In the second stage, we train an image encoder and a multi-camera fusion module built upon the pretrained decoder from the first stage. Specifically, for images from each camera, a shared image encoder predicts the peak direction vector $\mathbf{f}_{\rm dir}^{(i)}$, the intensity vector $\mathbf{f}_{\rm int}^{(i)}$, and the sky content vector $\mathbf{f}_{\rm content}^{(i)}$ for each image $\mathcal{I}^{(i)}$, where $i$ is the camera index. We design the latent vector fusion across the multiple camera views as follows: all $\mathbf{f}_{\rm dir}^{(i)}$ are aligned to the front-facing view using their extrinsic parameters and averaged to form $\bar{\mathbf{f}}_{\rm dir}$; all $\mathbf{f}_{\rm int}^{(i)}$ are averaged to yield $\bar{\mathbf{f}}_{\rm int}$; all $\mathbf{f}_{\rm content}^{(i)}$ are integrated into $\bar{\mathbf{f}}_{\rm content}$ through a self-attention module. Finally, the pretrained decoder reconstructs the HDR skydome image $\mathcal{I}_{\rm skydome}$ from $\bar{\mathbf{f}}_{\rm dir}$, $\bar{\mathbf{f}}_{\rm int}$ and $\bar{\mathbf{f}}_{\rm content}$. 

Compared to alternative approaches~\cite{wang2022neural,hold2019deep}, our multi-camera sky dome estimation technique accurately reproduces the sun's intensity response behavior at its peak with our residual designs, significantly improving the accuracy and fidelity of the skydome reconstruction. 

\begin{figure}[t] 
\centering
\includegraphics[width=0.42\textwidth]{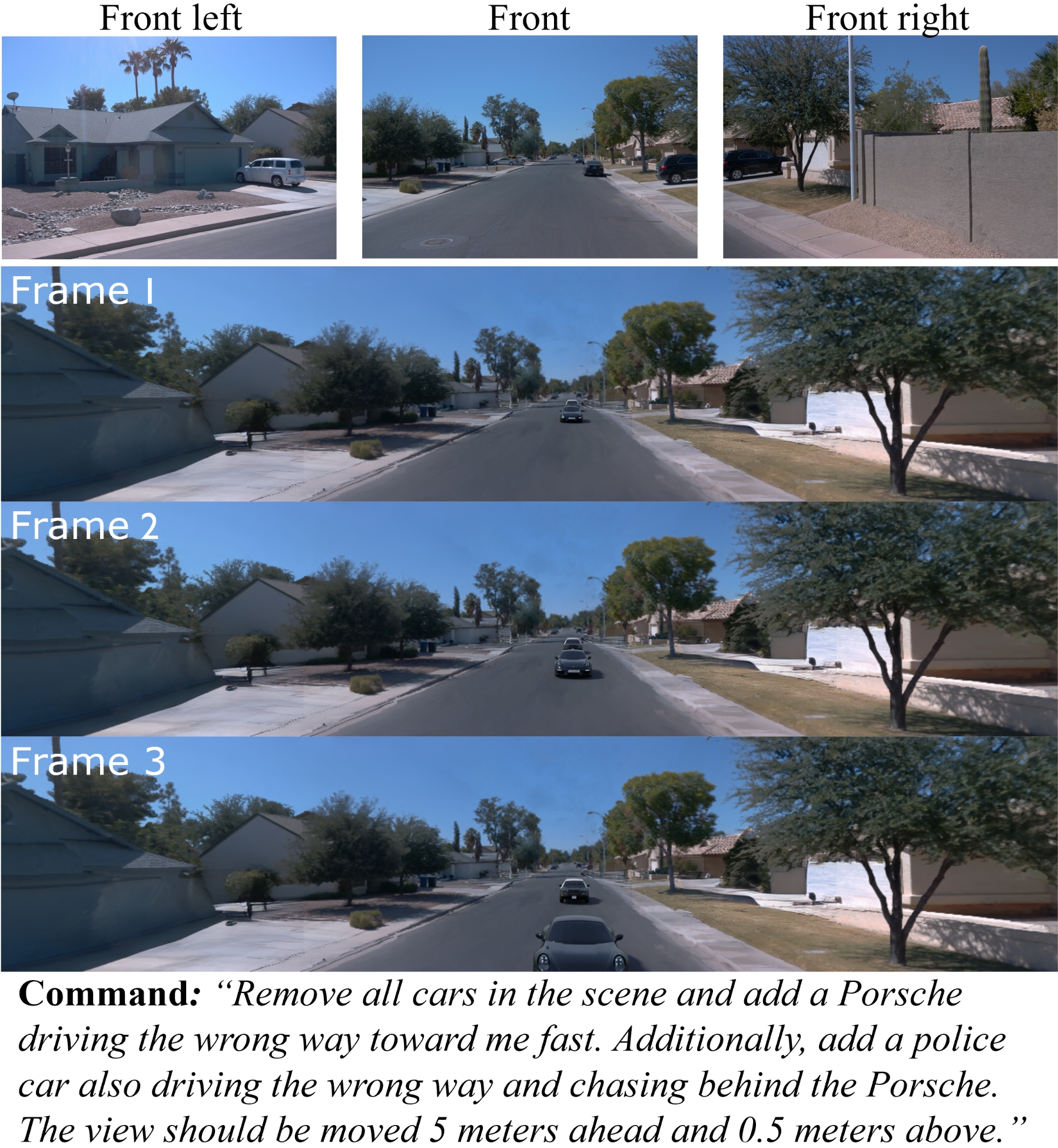}
\vspace{-3mm}
\caption{\small Editing result under a complex and mixed command.}
\label{fig:mix_result}
\vspace{-5mm}
\end{figure}

\noindent\textbf{Surrounding lighting estimation.}
Merely modeling the skydome cannot replicate the complex location-specific lighting effects, like those in the shade with sunlight blocked by trees or buildings. 
Our McNeRF is capable of storing precise 3D scene information, enabling us to capture the surrounding scene's impact on lighting. This approach facilitates the achievement of spatially-varying lighting estimation. Specifically, we sample the hemisphere rays at the virtual object's position $\textbf{o}$.
The rays' directions, $\textbf{d}_i, i=0, 1, \cdots, h\times w$, are aligned with pixel coordinates on a unit sphere using equirectangular projection from an environment map, where $h$ and $w$ are map's height and width. With the ray $\mathbf{r} = \mathbf{o} + t\mathbf{d}_i$, we query our McNeRF as Equation~\ref{eq:nerf} to obtain HDR surrounding lighting $\mathcal{I}_{\rm surround} (\textbf{o}, \textbf{d}_i)$. The surrounding lighting estimation reconstructs complex environmental lighting, achieving a spatially varying effect and high consistency with the background.


\noindent\textbf{Blending.}
We blend the HDR intensity value from the skydome and surrounding lighting by transmittance of the final sampling point from McNeRF. The idea is that the rays emitted outside the radiance fields will definitely hit the skydome. Given the direction $\textbf{d}_i$, we retrieve the skydome's intensity $\mathcal{I}_{\rm skydome}(\textbf{d}_i)$ with equirectangular projection. The final HDR light intensity $\mathcal{I}_{\rm env} (\textbf{o}, \textbf{d}_i)$ is a combination of scene and skydome: 
\begin{equation*}
\setlength\abovedisplayskip{2pt}
\setlength\belowdisplayskip{2pt}
\mathcal{I}_{\rm env}(\textbf{o}, \textbf{d}_i) = \mathcal{I}_{\rm surround} (\textbf{o}, \textbf{d}_i) + T_K \mathcal{I}_{\rm skydome} (\textbf{d}_i), 
\end{equation*}
where $T_K$ is the last sampling point's transmittance. 

McLight offers two main advantages: i) it explicitly recovers the illuminance behavior at the peak and use complementary information from multiple cameras to restore accurate skydome; and ii) it enables location-specific lighting with consideration of complex scene structures.

\begin{figure}[t] 
\centering
\includegraphics[width=0.42\textwidth]{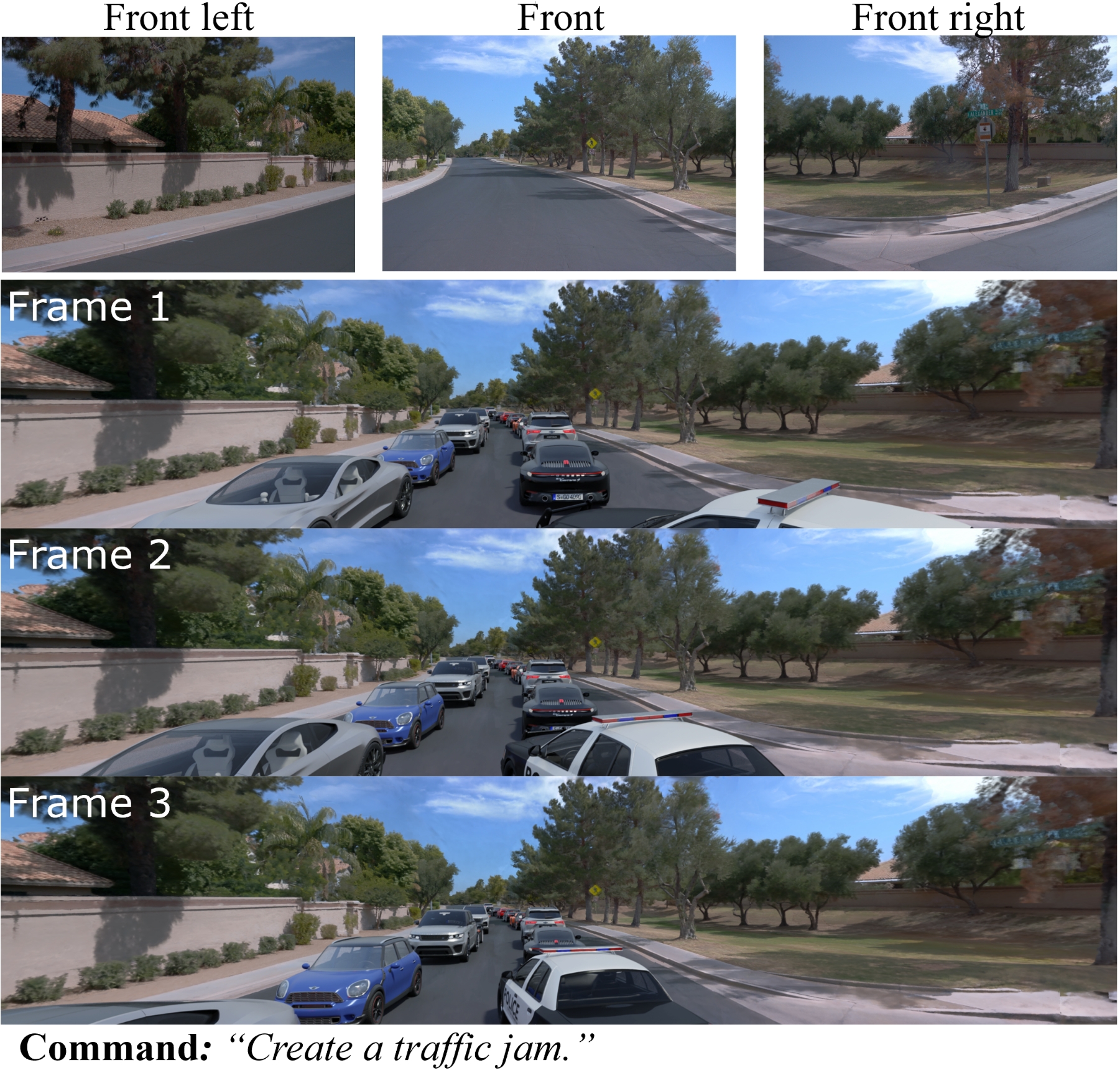}
\vspace{-3mm}
\caption{\small Editing result under a highly abstract command.}
\label{fig:abstract_result}
\vspace{-4mm}
\end{figure}
\section{Experimental Results}
\subsection{Datasets and implementation details}
\vspace{-1mm}
\quad
We demonstrate a variety of results mainly on the Waymo Open Dataset~\cite{Sun_2020_CVPR}, which contains high-quality multi-camera images and the corresponding calibrations. For McLight skydome estimation, we collect 449 HDRIs from online HDRI databases for the autoencoder training and use HoliCity \cite{zhou2020holicity}, a street view panorama dataset for the second stage; see more dataset details in the Appendix.

In our experiment, we use front, left front, and right front cameras in each frame. During the rendering process, we choose 40 frames per scene at a 10Hz sampling rate, totaling 120 images. We evenly select $1 / 8$ of these as the test set, with the remainder used for training. The input images are used at the dataset's initial resolution of $1920 \times 1280$; we employ GPT-4 as the LLMs in all of our experiments; see more implementation details in the Appendix.

\subsection{System results}
\label{subsec:system_results}
\textbf{Editing via language commands.} We select three representative commands to demonstrate the editing results. All of the results demonstrate we achieve photo-realistic wide angle results, thanks to McNeRF and McLight.

\emph{Mixed and complex command.} We send the system with a mixed and complex command, implying that a police car is chasing a wrong-way racer. The target scene, command and the result are shown in Figure \ref{fig:mix_result}. We see that i) every requirement in the complex command is accurately executed thanks to our multi-agent collaboration design; ii) this command successfully simulates one rare but dangerous driving condition, which is significant in accident testing. 

\emph{Highly abstract command.} The second type is a highly abstract command. The inputs and results are presented in Figure \ref{fig:abstract_result}. We see that i) this highly abstract command is hard to decompose by sentence division but still can be correctly executed by our method, and ii) our 3D asset bank offers a large variety of objects for addition.



\begin{figure}[t] 
\centering
\includegraphics[width=0.47\textwidth]{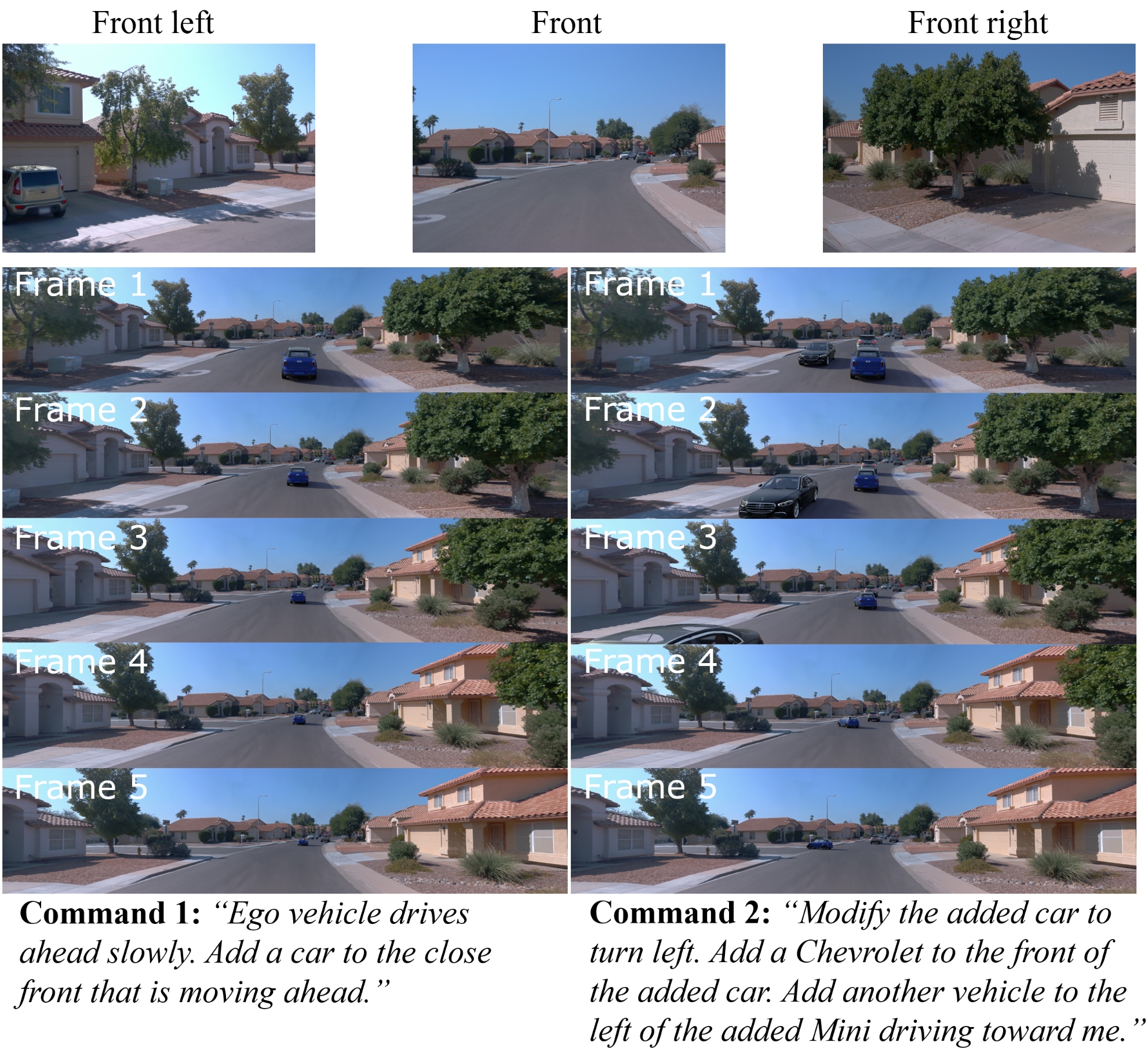}
\vspace{-3mm}
\caption{\small Editing result under multi-round commands.}
\label{fig:interactive_result}
\end{figure}

\emph{Multi-round command.} We also perform a multi-round chat with our system, and the commands in different rounds exist context dependencies. The final results are shown in Figure \ref{fig:interactive_result}. We see that i) our system is well-equipped to handle multi-round commands and execute the commands in each round precisely; ii) our system can handle the context dependencies across different rounds 
thanks to the recording ability of the project manager agent.

\begin{figure}[t] 
\centering
\includegraphics[width=0.45\textwidth]{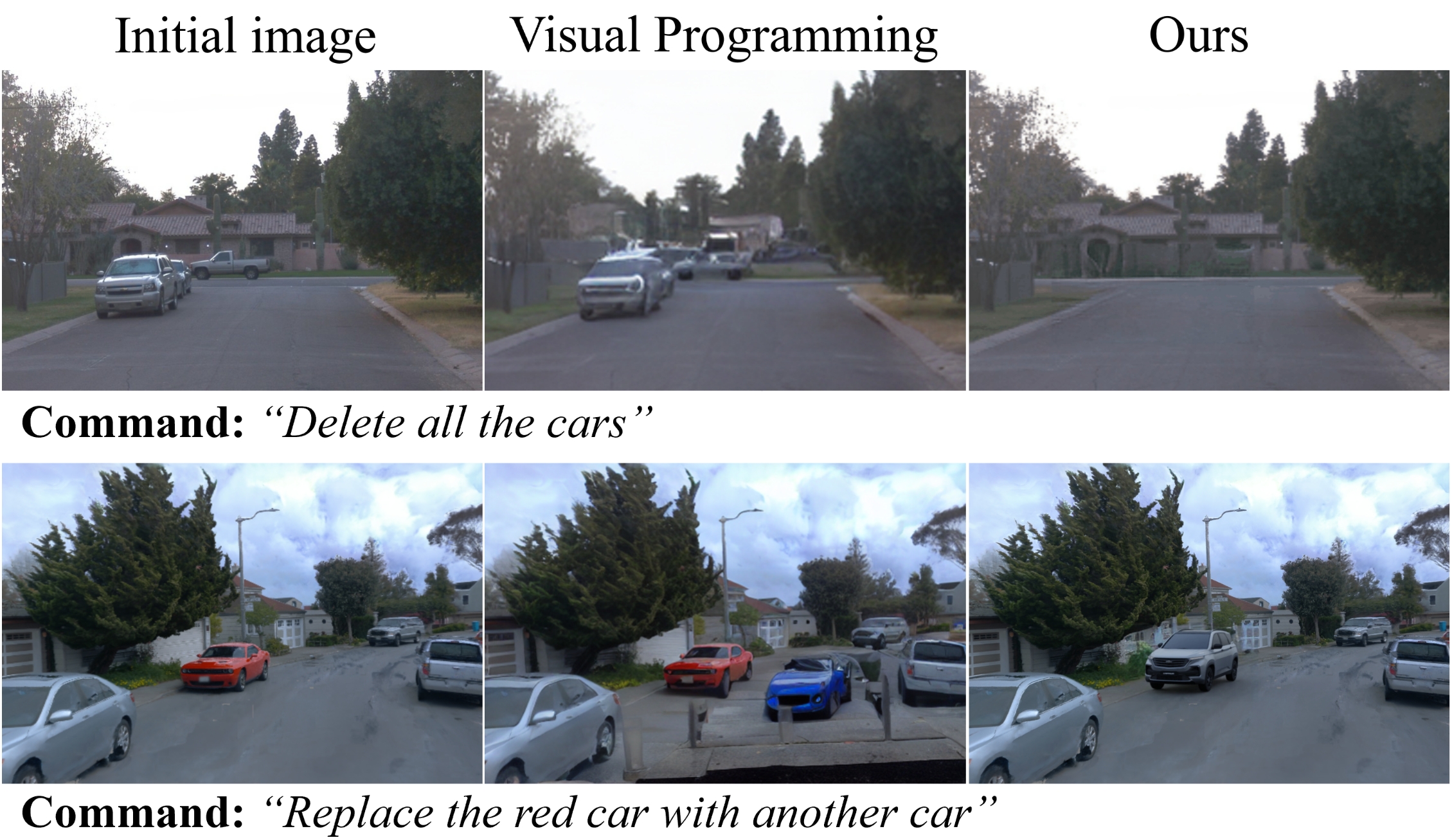}
\vspace{-3mm}
\caption{\small Qualitative comparison with Visual Programming\cite{gupta2023visual}. }
\label{fig:comparison_system_result}
\vspace{-4mm}
\end{figure}
\noindent\textbf{Comparison with Visual Programming~\cite{gupta2023visual}.}
Visual Programming (VP) is the latest SoTA language-driven 2D image neuro-symbolic system, which can also be used for editing with language commands. Given that Visual programming only supports editing of single frames in 2D images, we limit our comparison to the deletion and replacement operations within the original view that it can support. The comparison in Figure \ref{fig:comparison_system_result} shows that \textit{ChatSim} significantly outperforms VP in the two samples. Actually, in our extensive experiments, VP fails in most cases. The reason is that VP only has a single agent, making it difficult to handle mixed tasks. In comparison, \textit{ChatSim} has multiple collaborative agents with specific roles, ensuring accurate task execution.




\begin{figure}[t] 
\centering
\includegraphics[width=0.46\textwidth]{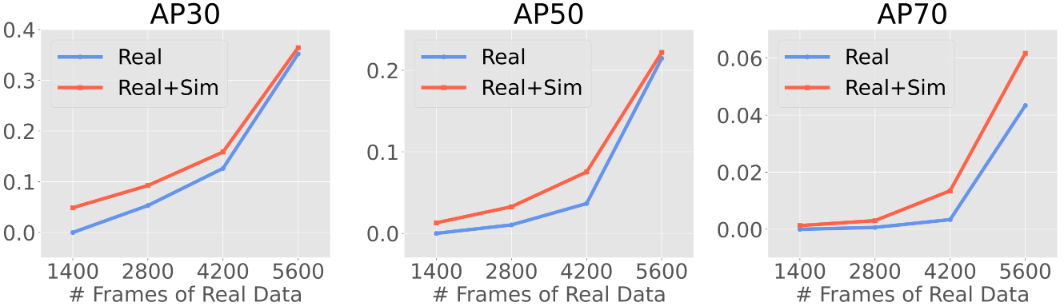}
\vspace{-3mm}
\caption{\small Comparison of detection performance w/o and with our simulated data under different amounts of real data during training. }
\label{fig:data_augment_comparison}
\end{figure}

\noindent\textbf{3D detection with simulation data.} 
We validate the benefits of our simulation as data augmentation for a
downstream 3D object detection task on Waymo Open Dataset~\cite{sun2020scalability}. We simulate 1960 frames, derived from scenes in the training dataset. In the simulation, cars with various types, locations, and orientations are incorporated. The detection model adopts Lift-Splat~\cite{philion2020lift}. Figure \ref{fig:data_augment_comparison} shows detection performances with and without fixed augmentation under various amounts of real data. We see that i) a significant and consistent improvement across different data sizes is achieved; ii) when real data is limited, our simulation notably aids in rough detection (AP30); iii) when the amount of real data increases, our simulation further significantly improves fine-grained detection (AP70), reflecting the high-quality of our simulation.

\subsection{Component results}
\vspace{-1mm}
\noindent\textbf{Multi-agent collaboration.}
We evaluated the effectiveness of the multi-agent collaboration by checking whether the command is successfully executed in Table \ref{tab:collaboration_exp}. In scenarios without multi-agent collaboration, all operations are executed by a single LLM agent. We see that a single LLM agent leads to notably lower execution accuracy across all categories due to process limitations. In contrast, the collaborative multi-agent approach can manage most commands, attributed to its task division and agent role specificity.

\begin{table}[t]
\scriptsize
\centering
\setlength\tabcolsep{4pt}
\resizebox{.93\columnwidth}{!}{
\begin{tabular}{c|ccccc}
\toprule
\multirow{2}*{\makecell{Multi-agent\\collaboration}}   &\multicolumn{5}{c}{Language command category}    \\
\cline{2-6}
~& {Deletion} &  {Addition} & {View change} & {Revision} & {Abstract} \\
\midrule
 &  0.617 & 0.383 & 0.717 & 0.367 & 0.216                                  \\
\checkmark    &  \textbf{0.983} &  \textbf{0.867} & \textbf{0.967} & \textbf{0.917} & \textbf{0.883}         \\
\bottomrule
\end{tabular}}
\vspace{-3mm}
\caption{\small The accuracy (\%) of task completion by LLM without and with multi-agent collaboration. }
\label{tab:collaboration_exp}
\vspace{-4mm}
\end{table}

\begin{table}[t]
\centering
\footnotesize
\resizebox{.93\columnwidth}{!}{
\begin{tabular}{c|cccc}
\toprule
Methods   & PSNR$\uparrow$ & SSIM$\uparrow$ & LPIPS$\downarrow$  & Inf. time (s)$\downarrow$ \\
\midrule
DVGO~\cite{sun2022direct}           &    23.57	&0.770	&0.508 & 7.7\\
Mip-NeRF360~\cite{barron2022mip}       &  24.40&	0.754&	0.528 & 101.8 \\
S-NeRF~\cite{xie2023s}           &    24.71  &   0.759   & 0.519 &   114.5   \\
F2NeRF~\cite{wang2023f2}          &     23.26	& 0.773&	0.439 & 2.4 \\
\midrule
Ours w/o alignment & 23.32 & 0.776 & 0.437 & 2.5\\
Ours w/o exposure & 25.18 & 0.819 & 0.381 & 2.4\\
\textbf{McNeRF (Ours)}  &   \textbf{25.82}	& \textbf{0.822} &	\textbf{0.378} &  2.5\\
\bottomrule
\end{tabular}
}
\vspace{-3mm}
\caption{\small Background novel view rendering performance evaluation. }
\label{render_exp}
\end{table}
\noindent\textbf{Background rendering.}
We compare our McNeRF with several other state-of-the-art methods on the background novel view synthesis task. We perform reconstruction and rendering on 32 selected scenes. Table~\ref{render_exp} shows the quantitative results comparison on three metrics: PSNR, SSIM, and LPIPS. We see that i) McNeRF achieves SoTA performance on all three metrics, significantly outperforming other baselines; ii) McNeRF has a fast inference speed, enabling quick responses to user requests for image rendering. 
\begin{figure}[t] 
\centering 
\includegraphics[width=0.5\textwidth]{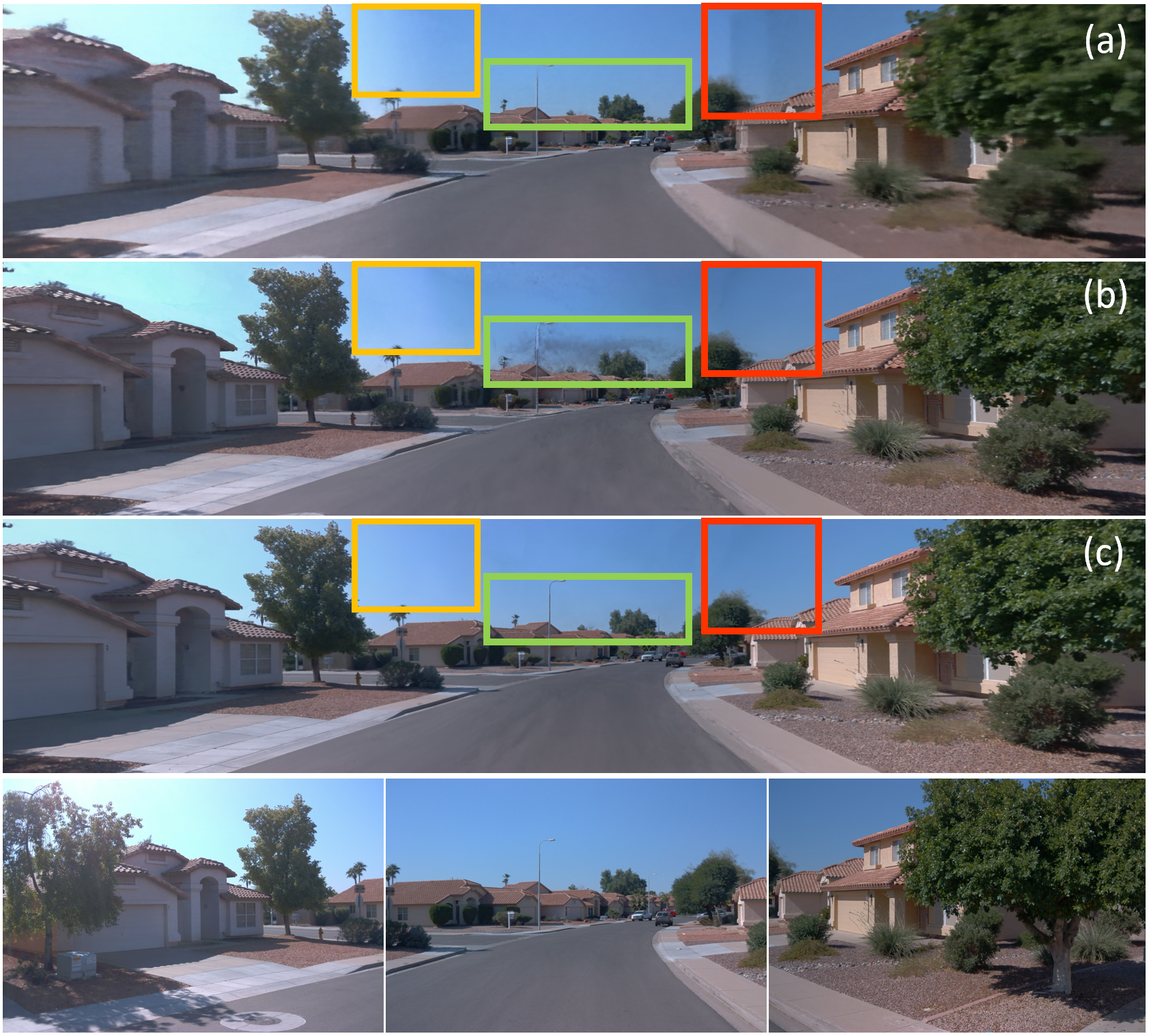} 
\vspace{-3mm}
\caption{\small Comparisons of wide-angle images generation. (a) S-NeRF.(b) F2NeRF. (c) McNeRF (Ours). Last row: target images.}
\label{Fig:nerf_comparision} 
\end{figure}

Figure~\ref{Fig:nerf_comparision} demonstrates qualitative comparisons between other methods and ours. We see that existing NeRF methods do not consider the exposure time, leading to noticeable changes in brightness at the junctions of different cameras in the image, as well as an overall inconsistency in exposure across the wide-angle view. Our method can make the brightness of the entire image more consistent.


\begin{table}[t]
\scriptsize
\setlength\tabcolsep{4pt}
\resizebox{.98\columnwidth}{!}{
\begin{tabular}{c|cc|cc|c}
\toprule
\multirow{2}*{Method} & \multicolumn{2}{c|}{Peak Intensity(log10) Error}  &\multicolumn{2}{c|}{Peak Angular Error (deg)} & \multirow{2}*{User study(\%) $\uparrow$}\\
~& Mean $\downarrow$  & Median $\downarrow$ & Mean $\downarrow$  & Median $\downarrow$ & ~\\
\midrule
Hold-Geoffroy et al. \cite{hold2019deep}                   & 0.899          & 0.975  & 48.4 & 51.6    &19.5     \\
Wang et al.  \cite{wang2022neural}                        & 0.590          & 0.628   &33.5&29.4    &37.3    \\
\textbf{McLight (Ours)}                           & \textbf{0.449} & \textbf{0.270} & \textbf{32.3} & \textbf{26.5} &\textbf{43.1}  \\ 
\bottomrule
\end{tabular}}
\vspace{-3mm}
\caption{\small Comparison with previous methods on lighting estimation.}
\label{tab:peak_int}
\vspace{-3mm}
\end{table}

\begin{figure*}[t]
    \centering
\includegraphics[width=0.97\textwidth]{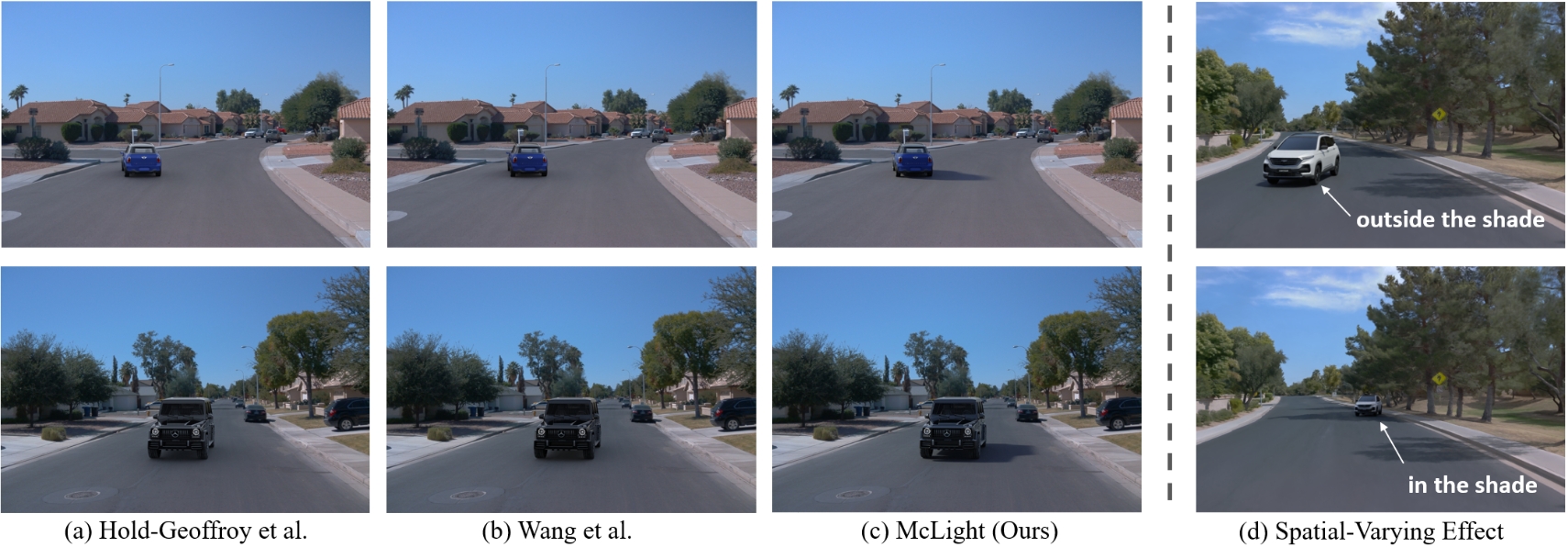}
    \vspace{-3mm}
    \caption{\small Comparison with different lighting estimation methods. 
    }
    \label{fig:vis_compare}
    \vspace{-3mm}
\end{figure*}

\noindent\textbf{Foreground rendering.}
We compare our McLight with the other two SoTA methods~\cite{wang2022neural,hold2019deep}. Table \ref{tab:peak_int} shows the comparison of relative intensity(log 10) error on our HDRI dataset, angular error on HoliCity~\cite{zhou2020holicity}, and user study. We see that McLight achieves more accurate peak behavior and receives noticeably higher user preferences. Figure \ref{fig:vis_compare} shows the visualizations of vehicle insertion. The vehicles added through McLight feature significantly more realistic reflections and strong shadows consistent with the scene.

\begin{table}[t]
\scriptsize
\centering
\resizebox{.98\columnwidth}{!}{
\begin{tabular}{c|ccccc}
\toprule
     Methods                     & Straight & Left Turn & Right Turn & Speed & Within-road \\
\hline
 GPT2Code  &  0.738   &   0.559   &   0.536   &   0.893  &  0.214  \\
 GPT2Motion&  0.595   &   0.119   &   0.167   &   0.345  &  0.277  \\
   \textbf{Ours}    & \textbf{0.988}   &   \textbf{0.940}   &  \textbf{0.976}   &   \textbf{0.952}  &  \textbf{1.000}  \\

\bottomrule
\end{tabular}}
\vspace{-3mm}
\caption{\small Comparison with motion generation from text methods.}
\label{tab:motion_exp}
\vspace{-6mm}
\end{table}

\noindent\textbf{Vehicle motion.}
As shown in Table \ref{tab:motion_exp}, we compare the motion generation method from user commands with two of our designed baselines:  1. GPT2Motion, which directly uses LLM to return the motion coordinates; 2. GPT2Code, which first generates code using LLM and executes it to obtain the vehicle motion. We validate multiple actions in multiple scenarios and report the user study result. The user study is to determine if the generated motions matched the command intents and fitted with the lane map. We see that our method demonstrated a significant advantage in generating motions from language commands. Additionally, it maintained a high rate of keeping the trajectories within the lane boundaries. 
\vspace{-2mm}
\section{Conclusions and Limitations}
\vspace{-1mm}
\quad
This paper introduces \textit{ChatSim}, the first system for editing 3D driving scene simulations via language commands 
and realistic rendering with import of external digital assets.
To effectively execute user commands, \textit{ChatSim} adopts an LLM-agent collaboration workflow. 
To promote photo-realistic simulation, we propose McNeRF and McLight for background and foreground rendering, respectively, accommodating multi-camera inputs. Experiments show that \textit{ChatSim} successfully simulates customized data via language commands, achieving high-quality, photo-realistic outcomes. \textbf{In future}, we plan to integrate more background editing functionalities to~\textit{ChatSim}, such as weather changes.
{
    \small
    \bibliographystyle{ieee_fullname}
    \bibliography{main}

\begin{thebibliography}{10}\itemsep=-1pt

\bibitem{51sim}
51Sim-One.
\newblock https://wdp.51aes.com/news/27.

\bibitem{agisoft2019metashape}
LLC Agisoft.
\newblock Metashape--photogrammetric processing of digital images and 3d spatial data generation, 2019.

\bibitem{akata2023playing}
Elif Akata, Lion Schulz, Julian Coda-Forno, Seong~Joon Oh, Matthias Bethge, and Eric Schulz.
\newblock Playing repeated games with large language models.
\newblock {\em arXiv preprint arXiv:2305.16867}, 2023.

\bibitem{amini2020learning}
Alexander Amini, Igor Gilitschenski, Jacob Phillips, Julia Moseyko, Rohan Banerjee, Sertac Karaman, and Daniela Rus.
\newblock Learning robust control policies for end-to-end autonomous driving from data-driven simulation.
\newblock {\em IEEE Robotics and Automation Letters}, 5(2):1143--1150, 2020.

\bibitem{anil2023palm}
Rohan Anil, Andrew~M Dai, Orhan Firat, Melvin Johnson, Dmitry Lepikhin, Alexandre Passos, Siamak Shakeri, Emanuel Taropa, Paige Bailey, Zhifeng Chen, et~al.
\newblock Palm 2 technical report.
\newblock {\em arXiv preprint arXiv:2305.10403}, 2023.

\bibitem{banta2016property}
Natalie~M Banta.
\newblock Property interests in digital assets: The rise of digital feudalism.
\newblock {\em Cardozo L. Rev.}, 38:1099, 2016.

\bibitem{barron2021mip}
Jonathan~T Barron, Ben Mildenhall, Matthew Tancik, Peter Hedman, Ricardo Martin-Brualla, and Pratul~P Srinivasan.
\newblock Mip-nerf: A multiscale representation for anti-aliasing neural radiance fields.
\newblock In {\em Proceedings of the IEEE/CVF International Conference on Computer Vision}, pages 5855--5864, 2021.

\bibitem{barron2022mip}
Jonathan~T Barron, Ben Mildenhall, Dor Verbin, Pratul~P Srinivasan, and Peter Hedman.
\newblock Mip-nerf 360: Unbounded anti-aliased neural radiance fields.
\newblock In {\em Proceedings of the IEEE/CVF Conference on Computer Vision and Pattern Recognition}, pages 5470--5479, 2022.

\bibitem{bergamini2021simnet}
Luca Bergamini, Yawei Ye, Oliver Scheel, Long Chen, Chih Hu, Luca Del~Pero, B{\l}a{\.z}ej Osi{\'n}ski, Hugo Grimmett, and Peter Ondruska.
\newblock Simnet: Learning reactive self-driving simulations from real-world observations.
\newblock In {\em 2021 IEEE International Conference on Robotics and Automation (ICRA)}, pages 5119--5125. IEEE, 2021.

\bibitem{boss2020two}
Mark Boss, Varun Jampani, Kihwan Kim, Hendrik Lensch, and Jan Kautz.
\newblock Two-shot spatially-varying brdf and shape estimation.
\newblock In {\em Proceedings of the IEEE/CVF Conference on Computer Vision and Pattern Recognition}, pages 3982--3991, 2020.

\bibitem{llm_gpt3}
Tom Brown, Benjamin Mann, Nick Ryder, Melanie Subbiah, Jared~D Kaplan, Prafulla Dhariwal, Arvind Neelakantan, Pranav Shyam, Girish Sastry, Amanda Askell, et~al.
\newblock Language models are few-shot learners.
\newblock {\em Advances in neural information processing systems}, 33:1877--1901, 2020.

\bibitem{caesar2020nuscenes}
Holger Caesar, Varun Bankiti, Alex~H Lang, Sourabh Vora, Venice~Erin Liong, Qiang Xu, Anush Krishnan, Yu Pan, Giancarlo Baldan, and Oscar Beijbom.
\newblock nuscenes: A multimodal dataset for autonomous driving.
\newblock In {\em Proceedings of the IEEE/CVF conference on computer vision and pattern recognition}, pages 11621--11631, 2020.

\bibitem{chang2019argoverse}
Ming-Fang Chang, John Lambert, Patsorn Sangkloy, Jagjeet Singh, Slawomir Bak, Andrew Hartnett, De Wang, Peter Carr, Simon Lucey, Deva Ramanan, et~al.
\newblock Argoverse: 3d tracking and forecasting with rich maps.
\newblock In {\em Proceedings of the IEEE/CVF conference on computer vision and pattern recognition}, pages 8748--8757, 2019.

\bibitem{chatgpt}
ChatGPT.
\newblock https://openai.com/blog/chatgpt.

\bibitem{chen2015deepdriving}
Chenyi Chen, Ari Seff, Alain Kornhauser, and Jianxiong Xiao.
\newblock Deepdriving: Learning affordance for direct perception in autonomous driving.
\newblock In {\em Proceedings of the IEEE international conference on computer vision}, pages 2722--2730, 2015.

\bibitem{chen2016monocular}
Xiaozhi Chen, Kaustav Kundu, Ziyu Zhang, Huimin Ma, Sanja Fidler, and Raquel Urtasun.
\newblock Monocular 3d object detection for autonomous driving.
\newblock In {\em Proceedings of the IEEE conference on computer vision and pattern recognition}, pages 2147--2156, 2016.

\bibitem{chen2017multi}
Xiaozhi Chen, Huimin Ma, Ji Wan, Bo Li, and Tian Xia.
\newblock Multi-view 3d object detection network for autonomous driving.
\newblock In {\em Proceedings of the IEEE conference on Computer Vision and Pattern Recognition}, pages 1907--1915, 2017.

\bibitem{chowdhery2022palm}
Aakanksha Chowdhery, Sharan Narang, Jacob Devlin, Maarten Bosma, Gaurav Mishra, Adam Roberts, Paul Barham, Hyung~Won Chung, Charles Sutton, Sebastian Gehrmann, et~al.
\newblock Palm: Scaling language modeling with pathways.
\newblock {\em arXiv preprint arXiv:2204.02311}, 2022.

\bibitem{international1999iec}
International~Electrotechnical Commission et~al.
\newblock Iec 61966-2-1: 1999 multimedia systems and equipment- colour measurement and management- part 2-1: Colour management- default rgb colour space- srgb, 1999.

\bibitem{blender}
Blender~Online Community.
\newblock {\em Blender - a 3D modelling and rendering package}.
\newblock Blender Foundation, Stichting Blender Foundation, Amsterdam, 2018.

\bibitem{crescenzi2001roadrunner}
Valter Crescenzi, Giansalvatore Mecca, Paolo Merialdo, et~al.
\newblock Roadrunner: Towards automatic data extraction from large web sites.
\newblock In {\em VLDB}, volume~1, pages 109--118, 2001.

\bibitem{dosovitskiy2017carla}
Alexey Dosovitskiy, German Ros, Felipe Codevilla, Antonio Lopez, and Vladlen Koltun.
\newblock Carla: An open urban driving simulator.
\newblock In {\em Conference on robot learning}, pages 1--16. PMLR, 2017.

\bibitem{du2023improving}
Yilun Du, Shuang Li, Antonio Torralba, Joshua~B Tenenbaum, and Igor Mordatch.
\newblock Improving factuality and reasoning in language models through multiagent debate.
\newblock {\em arXiv preprint arXiv:2305.14325}, 2023.

\bibitem{openscanerio}
Openscanerio Editor.
\newblock https://github.com/ebadi/openscenarioeditor.

\bibitem{UE}
Unreal Engine.
\newblock https://www.unrealengine.com/.

\bibitem{gao2023magicdrive}
Ruiyuan Gao, Kai Chen, Enze Xie, Lanqing Hong, Zhenguo Li, Dit-Yan Yeung, and Qiang Xu.
\newblock Magicdrive: Street view generation with diverse 3d geometry control.
\newblock {\em arXiv preprint arXiv:2310.02601}, 2023.

\bibitem{garon2019fast}
Mathieu Garon, Kalyan Sunkavalli, Sunil Hadap, Nathan Carr, and Jean-Fran{\c{c}}ois Lalonde.
\newblock Fast spatially-varying indoor lighting estimation.
\newblock In {\em Proceedings of the IEEE/CVF Conference on Computer Vision and Pattern Recognition}, pages 6908--6917, 2019.

\bibitem{guo2023streetsurf}
Jianfei Guo, Nianchen Deng, Xinyang Li, Yeqi Bai, Botian Shi, Chiyu Wang, Chenjing Ding, Dongliang Wang, and Yikang Li.
\newblock Streetsurf: Extending multi-view implicit surface reconstruction to street views.
\newblock {\em arXiv preprint arXiv:2306.04988}, 2023.

\bibitem{gupta2023visual}
Tanmay Gupta and Aniruddha Kembhavi.
\newblock Visual programming: Compositional visual reasoning without training.
\newblock In {\em Proceedings of the IEEE/CVF Conference on Computer Vision and Pattern Recognition}, pages 14953--14962, 2023.

\bibitem{hao2023chatllm}
Rui Hao, Linmei Hu, Weijian Qi, Qingliu Wu, Yirui Zhang, and Liqiang Nie.
\newblock Chatllm network: More brains, more intelligence.
\newblock {\em arXiv preprint arXiv:2304.12998}, 2023.

\bibitem{hold2019deep}
Yannick Hold-Geoffroy, Akshaya Athawale, and Jean-Fran{\c{c}}ois Lalonde.
\newblock Deep sky modeling for single image outdoor lighting estimation.
\newblock In {\em Proceedings of the IEEE/CVF conference on computer vision and pattern recognition}, pages 6927--6935, 2019.

\bibitem{hold2017deep}
Yannick Hold-Geoffroy, Kalyan Sunkavalli, Sunil Hadap, Emiliano Gambaretto, and Jean-Fran{\c{c}}ois Lalonde.
\newblock Deep outdoor illumination estimation.
\newblock In {\em Proceedings of the IEEE conference on computer vision and pattern recognition}, pages 7312--7321, 2017.

\bibitem{hong2023metagpt}
Sirui Hong, Xiawu Zheng, Jonathan Chen, Yuheng Cheng, Ceyao Zhang, Zili Wang, Steven Ka~Shing Yau, Zijuan Lin, Liyang Zhou, Chenyu Ran, et~al.
\newblock Metagpt: Meta programming for multi-agent collaborative framework.
\newblock {\em arXiv preprint arXiv:2308.00352}, 2023.

\bibitem{hu2023gaia}
Anthony Hu, Lloyd Russell, Hudson Yeo, Zak Murez, George Fedoseev, Alex Kendall, Jamie Shotton, and Gianluca Corrado.
\newblock Gaia-1: A generative world model for autonomous driving.
\newblock {\em arXiv preprint arXiv:2309.17080}, 2023.

\bibitem{kirillov2023segment}
Alexander Kirillov, Eric Mintun, Nikhila Ravi, Hanzi Mao, Chloe Rolland, Laura Gustafson, Tete Xiao, Spencer Whitehead, Alexander~C Berg, Wan-Yen Lo, et~al.
\newblock Segment anything.
\newblock {\em arXiv preprint arXiv:2304.02643}, 2023.

\bibitem{kundu2022panoptic}
Abhijit Kundu, Kyle Genova, Xiaoqi Yin, Alireza Fathi, Caroline Pantofaru, Leonidas~J Guibas, Andrea Tagliasacchi, Frank Dellaert, and Thomas Funkhouser.
\newblock Panoptic neural fields: A semantic object-aware neural scene representation.
\newblock In {\em Proceedings of the IEEE/CVF Conference on Computer Vision and Pattern Recognition}, pages 12871--12881, 2022.

\bibitem{lalonde2014lighting}
Jean-Fran{\c{c}}ois Lalonde and Iain Matthews.
\newblock Lighting estimation in outdoor image collections.
\newblock In {\em 2014 2nd international conference on 3D vision}, volume~1, pages 131--138. IEEE, 2014.

\bibitem{lalonde2010sun}
Jean-Fran{\c{c}}ois Lalonde, Srinivasa~G Narasimhan, and Alexei~A Efros.
\newblock What do the sun and the sky tell us about the camera?
\newblock {\em International Journal of Computer Vision}, 88:24--51, 2010.

\bibitem{legendre2019deeplight}
Chloe LeGendre, Wan-Chun Ma, Graham Fyffe, John Flynn, Laurent Charbonnel, Jay Busch, and Paul Debevec.
\newblock Deeplight: Learning illumination for unconstrained mobile mixed reality.
\newblock In {\em Proceedings of the IEEE/CVF Conference on Computer Vision and Pattern Recognition}, pages 5918--5928, 2019.

\bibitem{li2020rtm3d}
Peixuan Li, Huaici Zhao, Pengfei Liu, and Feidao Cao.
\newblock Rtm3d: Real-time monocular 3d detection from object keypoints for autonomous driving.
\newblock In {\em European Conference on Computer Vision}, pages 644--660. Springer, 2020.

\bibitem{li2023drivingdiffusion}
Xiaofan Li, Yifu Zhang, and Xiaoqing Ye.
\newblock Drivingdiffusion: Layout-guided multi-view driving scene video generation with latent diffusion model.
\newblock {\em arXiv preprint arXiv:2310.07771}, 2023.

\bibitem{li2023read}
Zhuopeng Li, Lu Li, and Jianke Zhu.
\newblock Read: Large-scale neural scene rendering for autonomous driving.
\newblock In {\em Proceedings of the AAAI Conference on Artificial Intelligence}, volume~37, pages 1522--1529, 2023.

\bibitem{li2018learning}
Zhengqin Li, Zexiang Xu, Ravi Ramamoorthi, Kalyan Sunkavalli, and Manmohan Chandraker.
\newblock Learning to reconstruct shape and spatially-varying reflectance from a single image.
\newblock {\em ACM Transactions on Graphics (TOG)}, 37(6):1--11, 2018.

\bibitem{liao2022maptr}
Bencheng Liao, Shaoyu Chen, Xinggang Wang, Tianheng Cheng, Qian Zhang, Wenyu Liu, and Chang Huang.
\newblock Maptr: Structured modeling and learning for online vectorized hd map construction.
\newblock {\em arXiv preprint arXiv:2208.14437}, 2022.

\bibitem{liao2023maptrv2}
Bencheng Liao, Shaoyu Chen, Yunchi Zhang, Bo Jiang, Qian Zhang, Wenyu Liu, Chang Huang, and Xinggang Wang.
\newblock Maptrv2: An end-to-end framework for online vectorized hd map construction.
\newblock {\em arXiv preprint arXiv:2308.05736}, 2023.

\bibitem{liu2023bevfusion}
Zhijian Liu, Haotian Tang, Alexander Amini, Xinyu Yang, Huizi Mao, Daniela~L Rus, and Song Han.
\newblock Bevfusion: Multi-task multi-sensor fusion with unified bird's-eye view representation.
\newblock In {\em 2023 IEEE International Conference on Robotics and Automation (ICRA)}, pages 2774--2781. IEEE, 2023.

\bibitem{mildenhall2021nerf}
Ben Mildenhall, Pratul~P Srinivasan, Matthew Tancik, Jonathan~T Barron, Ravi Ramamoorthi, and Ren Ng.
\newblock Nerf: Representing scenes as neural radiance fields for view synthesis.
\newblock {\em Communications of the ACM}, 65(1):99--106, 2021.

\bibitem{miric2019protecting}
Milan Miric, Kevin~J Boudreau, and Lars~Bo Jeppesen.
\newblock Protecting their digital assets: The use of formal \& informal appropriability strategies by app developers.
\newblock {\em Research Policy}, 48(8):103738, 2019.

\bibitem{muller2022instant}
Thomas M{\"u}ller, Alex Evans, Christoph Schied, and Alexander Keller.
\newblock Instant neural graphics primitives with a multiresolution hash encoding.
\newblock {\em ACM Transactions on Graphics (ToG)}, 41(4):1--15, 2022.

\bibitem{llm_gpt4}
OpenAI.
\newblock Gpt-4 technical report.
\newblock 2023.

\bibitem{ost2022neural}
Julian Ost, Issam Laradji, Alejandro Newell, Yuval Bahat, and Felix Heide.
\newblock Neural point light fields.
\newblock In {\em Proceedings of the IEEE/CVF Conference on Computer Vision and Pattern Recognition}, pages 18419--18429, 2022.

\bibitem{ost2021neural}
Julian Ost, Fahim Mannan, Nils Thuerey, Julian Knodt, and Felix Heide.
\newblock Neural scene graphs for dynamic scenes.
\newblock In {\em Proceedings of the IEEE/CVF Conference on Computer Vision and Pattern Recognition}, pages 2856--2865, 2021.

\bibitem{ouyang2022training}
Long Ouyang, Jeffrey Wu, Xu Jiang, Diogo Almeida, Carroll Wainwright, Pamela Mishkin, Chong Zhang, Sandhini Agarwal, Katarina Slama, Alex Ray, et~al.
\newblock Training language models to follow instructions with human feedback.
\newblock {\em Advances in Neural Information Processing Systems}, 35:27730--27744, 2022.

\bibitem{philion2020lift}
Jonah Philion and Sanja Fidler.
\newblock Lift, splat, shoot: Encoding images from arbitrary camera rigs by implicitly unprojecting to 3d.
\newblock In {\em Computer Vision--ECCV 2020: 16th European Conference, Glasgow, UK, August 23--28, 2020, Proceedings, Part XIV 16}, pages 194--210. Springer, 2020.

\bibitem{rombach2021highresolution}
Robin Rombach, Andreas Blattmann, Dominik Lorenz, Patrick Esser, and Björn Ommer.
\newblock High-resolution image synthesis with latent diffusion models, 2021.

\bibitem{rong2020lgsvl}
Guodong Rong, Byung~Hyun Shin, Hadi Tabatabaee, Qiang Lu, Steve Lemke, M{\=a}rti{\c{n}}{\v{s}} Mo{\v{z}}eiko, Eric Boise, Geehoon Uhm, Mark Gerow, Shalin Mehta, et~al.
\newblock Lgsvl simulator: A high fidelity simulator for autonomous driving.
\newblock In {\em 2020 IEEE 23rd International conference on intelligent transportation systems (ITSC)}, pages 1--6. IEEE, 2020.

\bibitem{sengupta2019neural}
Soumyadip Sengupta, Jinwei Gu, Kihwan Kim, Guilin Liu, David~W Jacobs, and Jan Kautz.
\newblock Neural inverse rendering of an indoor scene from a single image.
\newblock In {\em Proceedings of the IEEE/CVF International Conference on Computer Vision}, pages 8598--8607, 2019.

\bibitem{shah2018airsim}
Shital Shah, Debadeepta Dey, Chris Lovett, and Ashish Kapoor.
\newblock Airsim: High-fidelity visual and physical simulation for autonomous vehicles.
\newblock In {\em Field and Service Robotics: Results of the 11th International Conference}, pages 621--635. Springer, 2018.

\bibitem{somanath2021hdr}
Gowri Somanath and Daniel Kurz.
\newblock Hdr environment map estimation for real-time augmented reality.
\newblock In {\em Proceedings of the IEEE/CVF Conference on Computer Vision and Pattern Recognition}, pages 11298--11306, 2021.

\bibitem{sun2022direct}
Cheng Sun, Min Sun, and Hwann-Tzong Chen.
\newblock Direct voxel grid optimization: Super-fast convergence for radiance fields reconstruction.
\newblock In {\em Proceedings of the IEEE/CVF Conference on Computer Vision and Pattern Recognition}, pages 5459--5469, 2022.

\bibitem{sun2020scalability}
Pei Sun, Henrik Kretzschmar, Xerxes Dotiwalla, Aurelien Chouard, Vijaysai Patnaik, Paul Tsui, James Guo, Yin Zhou, Yuning Chai, Benjamin Caine, et~al.
\newblock Scalability in perception for autonomous driving: Waymo open dataset.
\newblock In {\em Proceedings of the IEEE/CVF conference on computer vision and pattern recognition}, pages 2446--2454, 2020.

\bibitem{Sun_2020_CVPR}
Pei Sun, Henrik Kretzschmar, Xerxes Dotiwalla, Aurelien Chouard, Vijaysai Patnaik, Paul Tsui, James Guo, Yin Zhou, Yuning Chai, Benjamin Caine, Vijay Vasudevan, Wei Han, Jiquan Ngiam, Hang Zhao, Aleksei Timofeev, Scott Ettinger, Maxim Krivokon, Amy Gao, Aditya Joshi, Yu Zhang, Jonathon Shlens, Zhifeng Chen, and Dragomir Anguelov.
\newblock Scalability in perception for autonomous driving: Waymo open dataset.
\newblock In {\em Proceedings of the IEEE/CVF Conference on Computer Vision and Pattern Recognition (CVPR)}, June 2020.

\bibitem{swerdlow2023street}
Alexander Swerdlow, Runsheng Xu, and Bolei Zhou.
\newblock Street-view image generation from a bird's-eye view layout.
\newblock {\em arXiv preprint arXiv:2301.04634}, 2023.

\bibitem{touvron2023llama}
Hugo Touvron, Louis Martin, Kevin Stone, Peter Albert, Amjad Almahairi, Yasmine Babaei, Nikolay Bashlykov, Soumya Batra, Prajjwal Bhargava, Shruti Bhosale, et~al.
\newblock Llama 2: Open foundation and fine-tuned chat models.
\newblock {\em arXiv preprint arXiv:2307.09288}, 2023.

\bibitem{turki2023suds}
Haithem Turki, Jason~Y Zhang, Francesco Ferroni, and Deva Ramanan.
\newblock Suds: Scalable urban dynamic scenes.
\newblock In {\em Proceedings of the IEEE/CVF Conference on Computer Vision and Pattern Recognition}, pages 12375--12385, 2023.

\bibitem{wang2023f2}
Peng Wang, Yuan Liu, Zhaoxi Chen, Lingjie Liu, Ziwei Liu, Taku Komura, Christian Theobalt, and Wenping Wang.
\newblock F2-nerf: Fast neural radiance field training with free camera trajectories.
\newblock In {\em Proceedings of the IEEE/CVF Conference on Computer Vision and Pattern Recognition}, pages 4150--4159, 2023.

\bibitem{wang2023drivedreamer}
Xiaofeng Wang, Zheng Zhu, Guan Huang, Xinze Chen, and Jiwen Lu.
\newblock Drivedreamer: Towards real-world-driven world models for autonomous driving.
\newblock {\em arXiv preprint arXiv:2309.09777}, 2023.

\bibitem{wang2022neural}
Zian Wang, Wenzheng Chen, David Acuna, Jan Kautz, and Sanja Fidler.
\newblock Neural light field estimation for street scenes with differentiable virtual object insertion.
\newblock In {\em European Conference on Computer Vision}, pages 380--397. Springer, 2022.

\bibitem{wang2023unleashing}
Zhenhailong Wang, Shaoguang Mao, Wenshan Wu, Tao Ge, Furu Wei, and Heng Ji.
\newblock Unleashing cognitive synergy in large language models: A task-solving agent through multi-persona self-collaboration.
\newblock {\em arXiv preprint arXiv:2307.05300}, 2023.

\bibitem{wu2017squeezedet}
Bichen Wu, Forrest Iandola, Peter~H Jin, and Kurt Keutzer.
\newblock Squeezedet: Unified, small, low power fully convolutional neural networks for real-time object detection for autonomous driving.
\newblock In {\em Proceedings of the IEEE conference on computer vision and pattern recognition workshops}, pages 129--137, 2017.

\bibitem{wu2023autogen}
Qingyun Wu, Gagan Bansal, Jieyu Zhang, Yiran Wu, Shaokun Zhang, Erkang Zhu, Beibin Li, Li Jiang, Xiaoyun Zhang, and Chi Wang.
\newblock Autogen: Enabling next-gen llm applications via multi-agent conversation framework.
\newblock {\em arXiv preprint arXiv:2308.08155}, 2023.

\bibitem{wu2023mars}
Zirui Wu, Tianyu Liu, Liyi Luo, Zhide Zhong, Jianteng Chen, Hongmin Xiao, Chao Hou, Haozhe Lou, Yuantao Chen, Runyi Yang, Yuxin Huang, Xiaoyu Ye, Zike Yan, Yongliang Shi, Yiyi Liao, and Hao Zhao.
\newblock Mars: An instance-aware, modular and realistic simulator for autonomous driving.
\newblock {\em CICAI}, 2023.

\bibitem{xie2023s}
Ziyang Xie, Junge Zhang, Wenye Li, Feihu Zhang, and Li Zhang.
\newblock S-nerf: Neural radiance fields for street views.
\newblock {\em arXiv preprint arXiv:2303.00749}, 2023.

\bibitem{xu2023drl}
Yinda Xu and Lidong Yu.
\newblock Drl-based trajectory tracking for motion-related modules in autonomous driving.
\newblock {\em arXiv preprint arXiv:2308.15991}, 2023.

\bibitem{yang2023bevcontrol}
Kairui Yang, Enhui Ma, Jibin Peng, Qing Guo, Di Lin, and Kaicheng Yu.
\newblock Bevcontrol: Accurately controlling street-view elements with multi-perspective consistency via bev sketch layout.
\newblock {\em arXiv preprint arXiv:2308.01661}, 2023.

\bibitem{yang2020surfelgan}
Zhenpei Yang, Yuning Chai, Dragomir Anguelov, Yin Zhou, Pei Sun, Dumitru Erhan, Sean Rafferty, and Henrik Kretzschmar.
\newblock Surfelgan: Synthesizing realistic sensor data for autonomous driving.
\newblock In {\em Proceedings of the IEEE/CVF Conference on Computer Vision and Pattern Recognition}, pages 11118--11127, 2020.

\bibitem{yang2023unisim}
Ze Yang, Yun Chen, Jingkang Wang, Sivabalan Manivasagam, Wei-Chiu Ma, Anqi~Joyce Yang, and Raquel Urtasun.
\newblock Unisim: A neural closed-loop sensor simulator.
\newblock In {\em Proceedings of the IEEE/CVF Conference on Computer Vision and Pattern Recognition}, pages 1389--1399, 2023.

\bibitem{zhang2019all}
Jinsong Zhang, Kalyan Sunkavalli, Yannick Hold-Geoffroy, Sunil Hadap, Jonathan Eisenman, and Jean-Fran{\c{c}}ois Lalonde.
\newblock All-weather deep outdoor lighting estimation.
\newblock In {\em Proceedings of the IEEE/CVF conference on Computer Vision and Pattern Recognition}, pages 10158--10166, 2019.

\bibitem{zhou2020holicity}
Yichao Zhou, Jingwei Huang, Xili Dai, Linjie Luo, Zhili Chen, and Yi Ma.
\newblock {HoliCity}: A city-scale data platform for learning holistic {3D} structures.
\newblock 2020.
\newblock arXiv:2008.03286 [cs.CV].

\bibitem{zhuge2023mindstorms}
Mingchen Zhuge, Haozhe Liu, Francesco Faccio, Dylan~R Ashley, R{\'o}bert Csord{\'a}s, Anand Gopalakrishnan, Abdullah Hamdi, Hasan Abed Al~Kader Hammoud, Vincent Herrmann, Kazuki Irie, et~al.
\newblock Mindstorms in natural language-based societies of mind.
\newblock {\em arXiv preprint arXiv:2305.17066}, 2023.

\end{thebibliography}
}

\clearpage
\appendix

\section{Supplementary Explanation of Table 2 }
Table 2 evaluates the execution success rate of commands of 5 instruction categories across 4 different driving sequences. For each category, the accuracy is measured as the average success rate across 3 trials of 15 commands that are specifically designed for this category. Each trial is deemed successfully executed if the LLM-agent(s) accurately perform the required operations, including setting correct configurations and parameter values.

\section{Supplementary Experiments}
\subsection{Supplementary Experiments of Lighting Estimation.}
We merge multi-view inputs into a wide-angle image for lighting estimation baselines with results in right table. McLight still significantly outperforms.  Intensity evaluation, not involving multi-view inputs, remains the same as paper Tab.4.
\begin{table}[h]
\scriptsize
\renewcommand\arraystretch{0.9}
\setlength\tabcolsep{4pt}
\resizebox{.98\columnwidth}{!}{
\begin{tabular}{c|cc|cc}
\toprule
Method, Multi-view (MV) version & MV Hold-Geoffroy & MV Wang  & \textbf{McLight (Ours)} \\
\midrule
Peak Angular Error(Mean/Median) & 36.7/37.1 & 33.7/29.3 & \textbf{32.3}/\textbf{26.5} \\
\hline
\end{tabular}}
\end{table}

\subsection{Supplementary Experiments of Background Rendering.}
Table below shows ablations to train baselines with the multi-camera alignment used in McNeRF. Multi-camera alignment is a general and practical trick which improves the rendering performance consistently, and our McNeRF(with exposure) still outperforms other baselines.
\begin{table}[h]
\renewcommand{\arraystretch}{0.9}
\centering
\scriptsize
\resizebox{.99\columnwidth}{!}{
\begin{tabular}{c|cccc}
\toprule
Methods    & PSNR$\uparrow$ & SSIM$\uparrow$ & LPIPS$\downarrow$  & Inf. time (s)$\downarrow$ \\
\midrule
DVGO + Alignment            &  24.65	&0.787	&0.487 & 7.7\\
Mip-NeRF360 + Alignment        & 25.50 &	0.759&	0.514 & 101.8 \\
S-NeRF + Alignment          &  25.53  &   0.760   & 0.513 &   114.5   \\
F2NeRF + Alignment          &   25.18	& 0.819&	0.381 & \textbf{2.4} \\
\textbf{F2NeRF + Alignment + Exposure (Ours)}   &  \textbf{25.82}	& \textbf{0.822} &	\textbf{0.378} &  2.5\\
\bottomrule
\end{tabular}
}

\end{table}

\section{LLM-Agents Details}
\subsection{Agent Implement Details}
LLM-Agents consist of their LLM (Large Language Model) component and corresponding functionalities. All experiments utilize the GPT-4 API\cite{llm_gpt4} to implement the LLM part. In each agent's prompt, there are elements involving the agent's function, the definition of actions that the agent needs to perform, the definition of information inputted to the agent, and the definition of outputs required from the agent. To facilitate the integration of Python code and ensure stable calls, the LLM part is required to return information in the format of a JSON dictionary. Additionally, each LLM part's prompt includes some examples, which contain inputs for certain scenarios and the corresponding expected outputs. If the input command does not contain the information of the keys of the output JSON dictionary, a default one will be filled in the dictionary. The parameters related to the GPT-4 API are all set to the official default values.

Note that, for supporting modification operations during multi-round commands, the 3D asset management agent, the vehicle motion agent, and the vehicle deleting agent have the ability to modify the information of already added or deleted cars.

\subsection{Reasoning Processes}
This section describes the natural language reasoning processes for the three cases presented in Section 5.2 of the main text.

\textbf{Mixed and complex command.} The initial input command is: \textit{"Remove all cars in the scene and add a Porsche driving the wrong way toward me fast. Additionally, add a police car also driving the wrong way and chasing behind the Porsche. The view should be moved 5 meters ahead and 0.5 meters above."} The command is decoupled by the project manager agent as following commands: 1.\textit{"Remove all cars."}; 2. \textit{"Add a Porsche driving the wrong way toward me fast."}; 3. \textit{"Add a police car also driving the wrong way and chasing behind the Porsche."}; 4. \textit{"The view should be moved 5 meters ahead and 0.5 meters above."}. The \textit{"Remove all cars."} command is distributed to the vehicle deleting agent, and then the agent finds the 3D boxes of all cars and applies the inpainting function for the removal operation. \textit{"Add a Porsche driving the wrong way toward me fast."} command is distributed to the 3D asset management agent for selecting the proper 3D asset. This command will also be distributed to the vehicle motion agent, which utilizes the key information in the command including "wrong way", "toward me" and "fast" to choose the appropriate start and end points and generate the motion with the motion generation function. \textit{"Add a police car also driving the wrong way and chasing behind the Porsche."} command will also be executed in the same way as the former operation. This command mentions the information of the added car, and the added car's information has been memorized by the project manager. This information is offered to the vehicle motion agent for determining the added police car's location. \textit{"The view should be moved 5 meters ahead and 0.5 meters above."} command is distributed to the view adjustment agent. The view adjustment agent returns the adjustment information of extrinsic as configuration, and calls the function to change the extrinsics to achieve view adjustment. Finally, background rendering and foreground rendering agents are required to generate the background and foreground results according to the information returned by the other agents, and the results are composed as the final outputs.

\textbf{Highly abstract command.} The initial input command is: \textit{"Create a traffic jam."} The project manager agent analyzes the command and decouples it as multiple repeats of car addition. These addition commands are processed by the 3D asset management agent and vehicle motion agent successively and are rendered by the foreground rendering agent. Combined with the rendered results from the background rendering agent, we can get the final outputs.

\textbf{Multi-round command.} The first initial command is: \textit{"Ego vehicle drives ahead slowly. Add a car to the close front that is moving ahead.”} The command is decoupled by the project manager agent as 1: \textit{"Ego vehicle drives ahead slowly."}; 2: \textit{"Add a car to the close front that is moving ahead."}. The first sub-command is distributed to view the adjustment agent, and the agent generates the extrinsics that represent moving ahead slowly. The second sub-command is executed as the process introduced above. 

The second initial command is: \textit{"Modify the added car to turn left. Add a Chevrolet to the front of the added car. Add another vehicle to the left of the added Mini driving toward me."} The command is decoupled by the project manager agent as 1: \textit{"Modify the added car to turn left."}; 2: \textit{"Add a Chevrolet to the front of the added car."}; 3: \textit{"Add another vehicle to the left of the added Mini driving toward me."} The first sub-command is distributed to the vehicle motion agent, which generates new motion based on the command for the determined added car. The following two sub-commands are executed in the same way as mentioned in the paragraphs above. Compositing the outputs of background rendering and foreground rendering agents can get the final outputs.

\section{Skydome Lighting Estimation Details}
\subsection{HDRI dataset}
We collect 449 high-quality outdoor panorama HDRIs from \href{https://polyhaven.com/hdris}{Poly Heaven Website}. These HDRIs are all licensed as CC0. We randomly selected 357 HDRIs for the training set and the remaining for the test set. A script for downloading these HDRIs will be available. 

\begin{figure*}[t]
    \centering
    \includegraphics[width=0.9\textwidth]{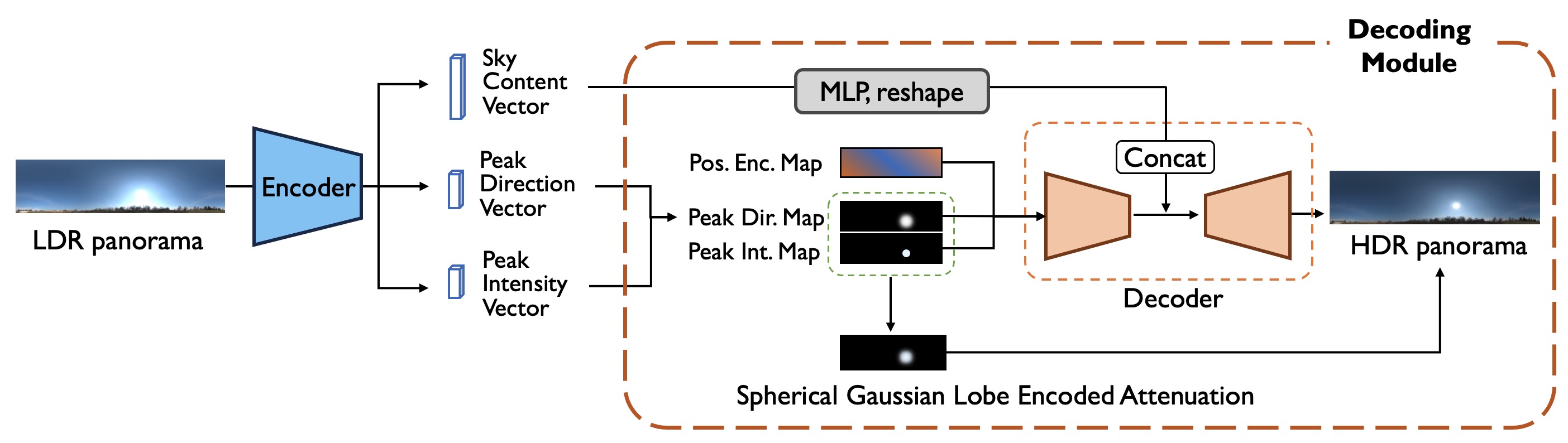}
        \vspace{-3mm}
    \caption{LDR to HDR reconstruction network. We add an explicit spherical Gaussian lobe encoded attenuation to overcome the over-smoothness in the decoded HDR panorama. It effectively ensures that the sun's intensity significantly exceeds that of surrounding pixels, rendering strong shadow effects for inserted objects. }
    \label{fig:ldr_to_hdr}
    \vspace{-3mm}
\end{figure*}

\begin{figure*}[t]
    \centering
    \includegraphics[width=0.9\textwidth]{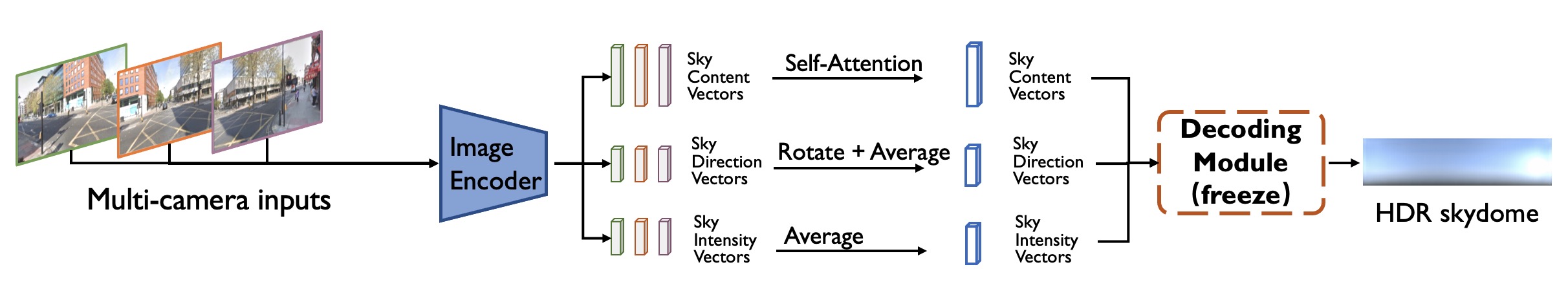}
    \vspace{-3mm}
    \caption{Reconstructing HDR skydome from multi-camera images. Training on HoliCity~\cite{zhou2020holicity} dataset.}
    \vspace{-3mm}
    \label{fig:predict_from_mc}
\end{figure*}

\subsection{LDR to HDR Skydome Reconstruction}
In this step, we utilize our HDRI dataset to train an LDR to HDR autoencoder with the aim of converting the skydome into a compact feature representation. We use the sRGB opto-electronic transfer function (also known as gamma correction) to get the LDR sky panorama, and follow ~\cite{wang2022neural} to transform the LDR sky panorama to 3 intermediate vectors, including the sky content vector $\mathbf{f}_{\rm content} \in \mathbb{R}^{64}$, the peak direction vector $\mathbf{f}_{\rm dir} \in \mathbb{R}^{3}$ and the intensity vector $\mathbf{f}_{\rm int} \in \mathbb{R}^{3}_{+}$. In the process of converting intermediate vectors into a reconstructed HDR sky panorama, we construct the peak direction map \(\mathbf{M}_{\rm dir}\), the peak intensity map \(\mathbf{M}_{\rm int}\) and the positional encoding map \(\mathbf{M}_{\rm pe}\).

Peak direction map (\(\mathbf{M}_{\rm dir}\)): For each pixel in \(\mathbf{M}_{\rm dir}\), we calculate the peak direction embedding. This calculation utilizes a spherical Gaussian lobe, formulated as \(\mathbf{M}_{\rm dir}(\mathbf{u}) = e^{100*(\mathbf{u}\cdot \mathbf{f}_{dir} -1)}\), where \(\mathbf{f}_{\rm dir}\) denotes the peak direction vector. This map is represented in  \(\mathbb{R}^{H \times W \times 1}\).

Peak intensity map (\(\mathbf{M}_{\rm int}\)): Each pixel in this map is determined based on its corresponding value in the peak direction map. Specifically, for a given direction \(\mathbf{u}\), if \(\mathbf{M}_{\rm dir}(\mathbf{u}) > 0.9\), then \(\mathbf{M}_{\rm int}(\mathbf{u})\) is assigned the value of \(\mathbf{f}_{\rm int}\). If not, \(\mathbf{M}_{\rm int}(\mathbf{u})\) is set to zero. This map is represented in  \(\mathbb{R}_{+}^{H \times W \times 3}\).

Positional encoding map (\(\mathbf{M}_{\rm pe}\)): This map encodes the direction vector of each pixel, determined through equirectangular projection, thus contributing to the accurate reconstruction of the HDR sky panorama. It is defined in  \(\mathbb{R}^{H \times W \times 3}\).

The input of the decoder $\mathbf{M}_{\rm input}$ is a concatenation of $\mathbf{M}_{\rm pe}, \mathbf{M}_{\rm dir}$ and $ \mathbf{M}_{\rm int}$. We use a 2D UNet to decode the concatenated input map to the HDR sky panorama. For sky content vector $\mathbf{f}_{\rm content}$, we use an MLP to increase its feature dimension, reshape it to a 2D feature map, and concatenate it with the intermediate features at the bottleneck of the UNet. This concatenated feature will be further decoded to the HDR sky panorama.

In the context of HDR imaging, the intensity of the peak often exhibits characteristics akin to an impulse response, displaying pixel values that are significantly elevated by orders of magnitude in comparison to adjacent pixels. This presents a substantial challenge for the decoder in accurately recovering these patterns. Thus, we design a residual connection to explicitly inject the peak intensity information into the final HDR sky panorama. Let \(\mathbf{M}_{\rm peak}\) be the product of \(\mathbf{M}_{\rm dir}\) and \(\mathbf{M}_{\rm int}\), representing an attenuation encoded by a spherical Gaussian lobe. In our design, we specifically substitute the decoded HDR sky panorama at the peak position with \(\mathbf{M}_{\rm peak}\). This substitution is applied where the value of \(\mathbf{M}_{\rm int}(\mathbf{u})\) is non-zero, ensuring that the peak position in the HDR sky panorama is accurately represented by \(\mathbf{M}_{\rm peak}\). This makes a significant difference between us and ~\cite{wang2022neural}. Accurate and strong peak intensity can generate very strong shadow effects, resulting in better rendering realism. See Figure ~\ref{fig:ldr_to_hdr}.

To train the LDR to HDR skydome reconstruction, we computer the ground truth peak direction \(\mathbf{f}_{\rm dir}^{\rm gt}\) and peak intensity \(\mathbf{f}_{\rm int}^{\rm gt}\) from the HDR ground-truth. During the network training process, we employ four losses for supervision. These losses are as follows: 
peak direction loss $L_{\rm dir}$, which measures the L1 angular error of the peak direction vectors;
peak intensity loss $L_{\rm int}$, which quantifies the log-encoded L2 error of the peak intensity vectors;
HDR reconstruction loss $L_{\rm hdr-recon}$, which evaluates the log-encoded L2 error between the reconstructed HDR output and the ground truth HDR data;
LDR reconstruction loss $L_{\rm ldr-recon}$, which is calculated as the L1 error between the input LDR sky panorama and the gamma-corrected HDR reconstruction. 

The total loss is $L_{\rm total} = \lambda_1 L_{\rm dir} + \lambda_2 L_{\rm int} + \lambda_3 L_{\rm hdr-recon} + \lambda_4 L_{\rm ldr-recon}$, where $\lambda_1=1, \lambda_2=0.1, \lambda_3=2$ and $\lambda_4=0.2$. 

Data augmentation methods, including rotation, flipping, exposure adjustment and white balance adjustment, are implemented to enrich the training data. Noticing a strong white balance inaccuracy (the color temperature is too high) in the image data from Waymo Open Dataset~\cite{sun2020scalability}, we augment the HDRI with corresponding white balance adjustment. The blue channel is randomly enlarged by 1.2-1.3 times, and the red channel is randomly reduced by 1.2-1.3 times.

\subsection{Predict HDR Skydome from Multi-Camera Images}
This step involves estimating skydome lighting from multi-camera images collected by the vehicle. The core idea is to estimate intermediate features from multiple views and restore the skydome lighting using the well-trained HDR reconstruction decoding module. We emphasize the fusion of intermediate features from multiple cameras to get a complementary and comprehensive prediction for the skydome lighting.

Multi-camera image data will first go through a shared image encoder to predict the peak direction vector $\mathbf{f}_{\rm dir}^{(i)}$, the intensity vector $\mathbf{f}_{\rm int}^{(i)}$, and the sky content vector $\mathbf{f}_{\rm content}^{(i)}$ for each image $\mathcal{I}^{(i)}$, where $i$ is the camera index. For those vectors from $N$ cameras, we fuse all the features in the following strategy:

 We transform $\mathbf{f}_{\rm dir}^{(i)}, i=1,2,...,N$ to the front-facing view using their extrinsic parameters and averaged the rotated direction vector to $\bar{\mathbf{f}}_{\rm dir}$; we average $\mathbf{f}_{\rm int}^{(i)}, i=1,2,...,N$ to $\bar{\mathbf{f}}_{\rm int}$; we utilize the attention mechanism to fuse sky content vectors as $\bar{\mathbf{f}}_{\rm content} = \texttt{Attn(q, k, v)}$, where $\texttt{q} = \mathbf{f}^{(0)}_{\rm content}$, $\texttt{k} =\texttt{v} = \texttt{stack} (\{\mathbf{f}^{i}_{\rm content}\}_{i=0,1,...,N-1})$. Here index 0 refers to the first (front-facing) view image and $\texttt{Attn}(\cdot,\cdot,\cdot)$ the standard attention operator. Given $\bar{\mathbf{f}}_{\rm dir}$, $\bar{\mathbf{f}}_{\rm int}$, $\bar{\mathbf{f}}_{\rm content}$, we use the pre-trained decoding module from the previous stage to recover the fused intermediate vectors to HDR panorama. See Figure ~\ref{fig:predict_from_mc}.

Since there is no relevant panoramic data in the autonomous driving dataset for supervision, We use HoliCity~\cite{zhou2020holicity} to simulate multi-camera images.  Based on the arrangement and FOV of the three forward-facing cameras on the Waymo vehicle~\cite{sun2020scalability}, we cropped the corresponding image from the HoliCity panorama as the model inputs. To supervise the learning of the image encoder, we use the LDR to HDR reconstruction network from the previous stage to generate pseudo peak intensity vector GT, peak direction vector GT, sky content vector GT, and HDR skydome GT. 

We apply five losses to supervise the network during training. These losses are as follows: the peak direction loss $L_{\rm dir}$, which measures the L1 angular error of the fused peak direction vector; the peak intensity loss $L_{\rm int}$, which calculates the log-encoded L2 error of the fused peak intensity vectors; the sky content loss $L_{\rm content}$, which evaluates the L1 error of the fused sky content vectors; the HDR reconstruction loss $L_{\rm hdr-recon}$ with log-encoded L2 error; the LDR reconstruction $L_{\rm ldr-recon}$ with L1 error. 

The total loss is $L_{\rm total} = \lambda_1 L_{\rm dir} + \lambda_2 L_{\rm int} + \lambda_3 L_{\rm content} + \lambda_4 L_{\rm hdr-recon} + \lambda_5 L_{\rm ldr-recon}$, where $\lambda_1=0.5, \lambda_2=0.25, \lambda_3=0.005, \lambda_4=0.1$ and $\lambda_5=0.2$.

\section{3D Asset Bank}
To ensure ease of access and modification of 3D assets, we normalize our Blender models within their Blender files using the following procedure:

\begin{enumerate}
    \item We ensure that the model has accurate physical dimensions in the unit of meter.
    
    \item The origin of the car model is set at the middle of the bottom of the car. We position the model at the center of the world coordinate system, ensuring that the car model's origin aligns with the origin of the world coordinate system. The car is oriented to face the positive direction of the x-axis.

    \item We uniformly apply the \texttt{Principled BSDF} material to the car body, and name the material "car\_paint". Prompt that changes the asset's color will affect the "Base Color" attribute of the \texttt{Principled BSDF} node.

    \item We use the \texttt{Join} operator to merge all meshes into one object. 

\end{enumerate}

Following the aforementioned approach, we normalize the Blender models collected from the Internet to continuously expand our 3D Asset Bank.

\section{Blender Rendering Details}
We fully implement the Blender rendering workflow using Python scripting, incorporating features such as alpha channel, depth channel, and shadow effect, \textbf{all achieved within a single rendering pass}.

\begin{enumerate}
    \item To get a transparent background, we first enable the \texttt{Render Properties - Film - Transparent} option.
    \item To get multiple rendering output, we enable the \texttt{Combined} pass, \texttt{Z} pass and \texttt{Shadow Catcher} pass in \texttt{View Layer Properties} panel.
    \item To render the shadow, we add a very large plane under the car and enable the plane's \texttt{Object Properties - Visibility - Mask - Shadow Catcher} option. 
    \item To obtain scene-related colored shadows, we construct the compositing node graph as Figure ~\ref{fig:blender}. This configuration generates the rendered image overlaid on the scene image, along with the accompanying depth information and mask of the vehicle and its corresponding shadow.
    \begin{figure}[t]
        \centering
        \includegraphics[width=\columnwidth]{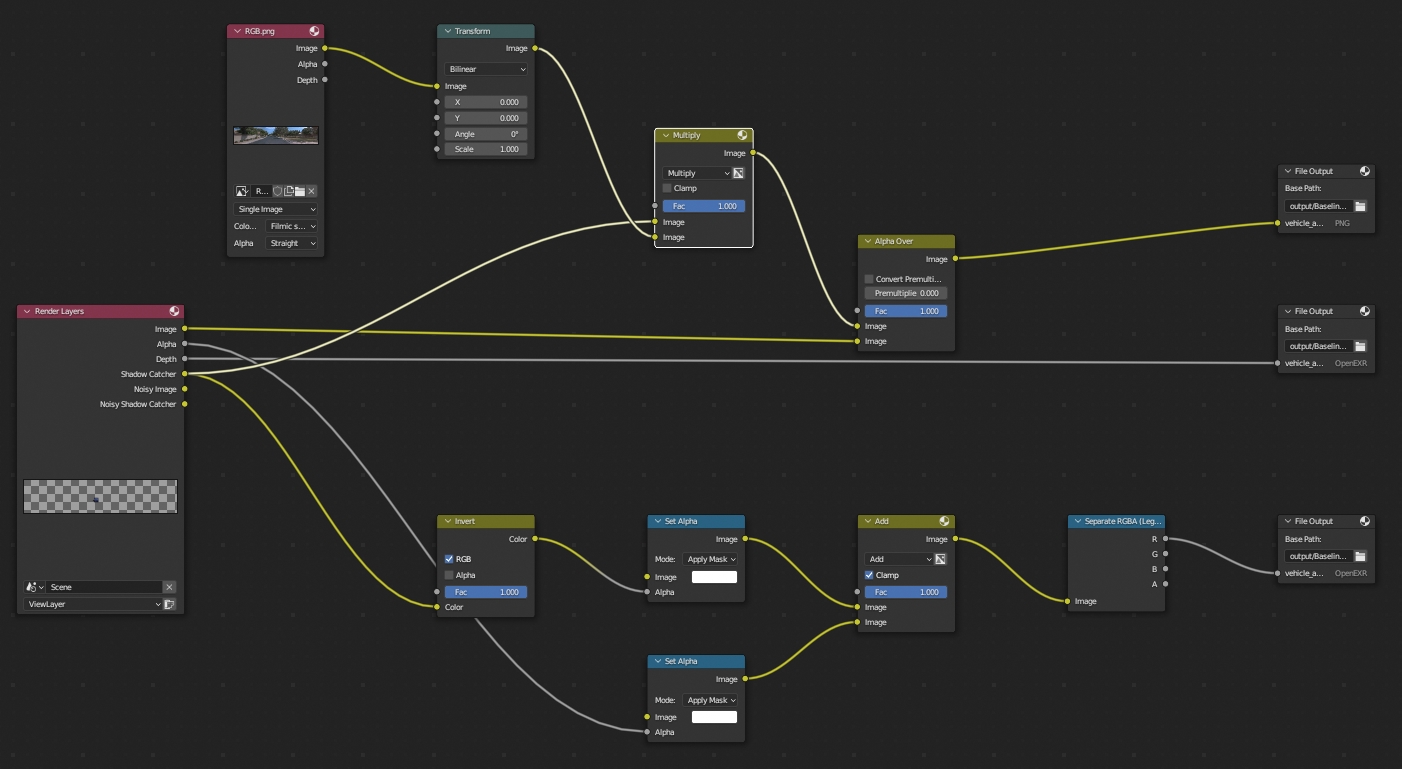}
            \vspace{-3mm}
        \caption{Compositing node graph design in Blender~\cite{blender}}
        \label{fig:blender}
            \vspace{-3mm}
    \end{figure}
    \item Using depth information and mask, we can handle the occlusion relationship with the original objects in the scene. We also added a moderate amount of motion blur to the rendered car to match the background.
\end{enumerate}

\section{Motion Generation Details}
The vehicle motion agent creates the initial places and subsequent motions of vehicles following the requests commands. Existing vehicle motion generation methods cannot directly generate motion purely from text and the scene map. Here we elaborate on the details of our text-to-motion methods. Our method consists of two parts: 
vehicle placement to generate the starting points and vehicle motion planning to generate the subsequent motions.

\begin{figure}
    \centering
    \includegraphics[width=0.38\textwidth]{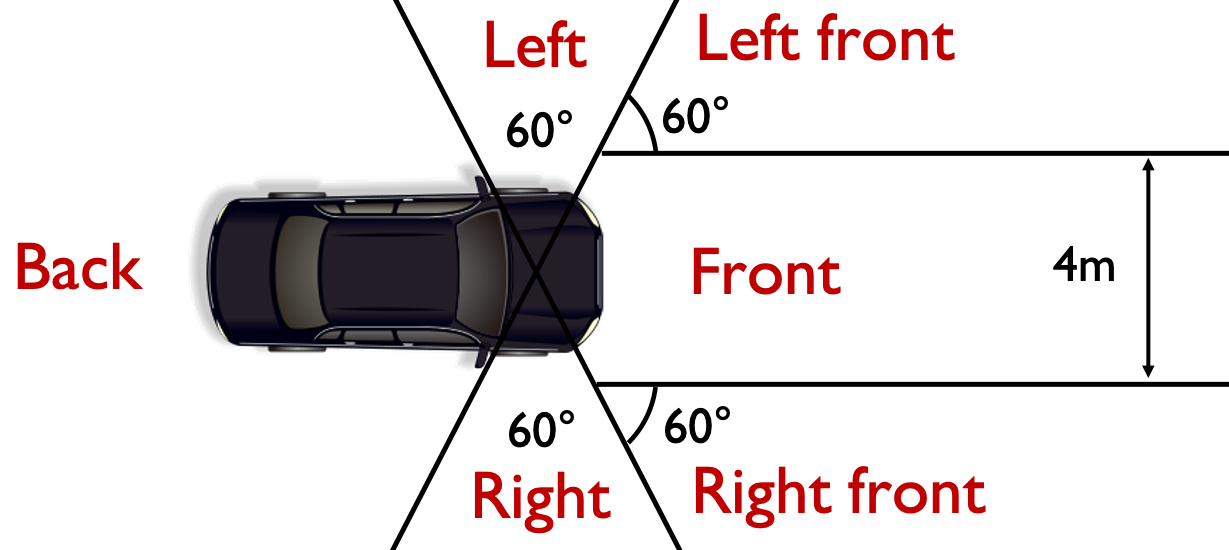}
    \caption{The neighboring area division for vehicle placement.}
    \label{fig:vehicle_placement}
    \vspace{-3mm}
\end{figure}

\subsection{Vehicle Placement}
We use the language command and the scene map to generate the initial position. The scene map $\mathcal{M}$ follows the lane map form $\mathcal{M}=\{\mathbf{n}_i,i=1,2,\cdots,m\}$, where $m$ is the number of lane nodes and the $i$th lane node $\mathbf{n}_i=(x_{\rm s},y_{\rm s},x_{\rm e},y_{\rm e},c_{\rm type})$ consists of lane starting position $(x_{\rm s},y_{\rm s})$, ending position $(x_{\rm e},y_{\rm e})$ and the lane type $c_{\rm type}$. The map range is cropped with the range of front 80m, left 20m and right 20m. Generally, we use the lane map from the ground-truth data. If the lane map does not exist, it is applicable to use a lane map estimation method like \cite{liao2022maptr,liao2023maptrv2} to obtain the lane map. 

Given the language command, the LLM first extracts key placement attributes, including vehicle number, distance range, relative direction with the observer and direction of driving, and crazy mode. With these attributes, the role function of placement begins to find suitable lane nodes from the scene map. Here we assume all the placed vehicles are on the centerline of the road. If the distance range $(d_{\rm min},d_{\rm max})$ is identified, the role function selects the lane centerline nodes according to their distance with the ego location. For the relative direction, we divide the ego neighboring area into 6 categories: front, left front, right front, left, right, and back, see Figure~\ref{fig:vehicle_placement} for illustration. For the direction of driving, we consider two types: driving close to the ego and driving away from the ego, which determines the left/right side of the vehicle on the road. The crazy mode, which is designed for non-compliant inverse driving behavior, is a bool variable. When it is true, we will inverse the direction of the map (swap the starting and ending point of each lane) for that vehicle to represent inverse driving. We select the matched lane node set and randomly select one lane node from the set. We also consider the conflict of placing vehicles by an iterative approach that incoming vehicles should not overlap with the existing vehicles. After obtaining lane nodes for every vehicle, we set the midpoint of the lane node to be the initial position of a vehicle and the direction of the lane to be the initial heading of the vehicle.

\subsection{Vehicle Motion Planning}
After obtaining the initial positions, we generate motions in two steps: plan the destination and plan the middle trajectory. We first extract movement attributes including speed, action, interval and time length. Notably, we divide actions into 5 categories: straightforward, turn left, turn right, park, and backward.  
To obtain the destination, if the action category is straightforward or park, and backward, we directly calculate a raw destination by assuming the car driving following a line with the target speed. Then we find the closest lane node with the raw destination to be the final destination.
If the action category is turning left or turning right, we select a set of nodes whose vertical distance with the initial line of heading is in a range (5m-30m) and fit the driving directions (the direction of the line should be away from the starting point). We randomly pick a lane node to be the destination. 

To plan the middle trajectory, we use an iterative adjustment approach to make the trajectory match with the map and avoid off-road driving. We first use one cubic Bezier curve to fit the overall trajectory with the condition of starting point, starting direction, ending point and ending direction. The cubic Bezier curve is formulated by
\begin{equation}
\begin{aligned}
    B(t) = & (1-t)^3P_0 + 3t(1-t)^2P_1 \\
    &+ 3t^2(1-t)P_2 + t^3P_3, t\in [0,1], 
\end{aligned}
\end{equation}
where $P_0,P_1,P_2,P_3 \in \mathbb{R}^2$ is the control points that can be solved by given starting point, starting direction, ending point and ending direction. Then to avoid off-road driving of the intermediate trajectory, we adjust the middle coordinate by replacing it with the closest lane node. We split the whole trajectory into two parts with the boundary of the middle coordinate and use one cubic Bezier curve to fit each split trajectory. We iteratively repeat the process to represent the planned trajectory by multiple cubic Bezier curves. Finally, to make the planned trajectory fit with vehicle dynamics, we use a trajectory tracking method in \cite{xu2023drl} as post-processing to revise the planned trajectory\footnote{\href{https://drl-based-trajectory-tracking.readthedocs.io/en/latest/}{https://drl-based-trajectory-tracking.readthedocs.io/en/latest/}}.

\section{Background rendering details}
\subsection{Dataset Selection}
For all Waymo Open Dataset~\cite{sun2020scalability} experiments, we use images captured from three frontal cameras. The details of selection are shown in Table~\ref{dataset}. There are 120 images in total for each scenerio.

\begin{table}[]
\centering
\scriptsize
\begin{tabular}{c|c}
\toprule
Sequence & Start Frame \\
\midrule
segment-10247954040621004675\_2180\_000\_2200\_000 & 0 \\
segment-13469905891836363794\_4429\_660\_4449\_660 & 40 \\
segment-14333744981238305769\_5658\_260\_5678\_260 & 40 \\
segment-1172406780360799916\_1660\_000\_1680\_000 & 50 \\
segment-4058410353286511411\_3980\_000\_4000\_000 & 90 \\
segment-10061305430875486848\_1080\_000\_1100\_000 & 30 \\
segment-14869732972903148657\_2420\_000\_2440\_000 & 0 \\
segment-16646360389507147817\_3320\_000\_3340\_000 & 0 \\
segment-13238419657658219864\_4630\_850\_4650\_850 & 0 \\
segment-14424804287031718399\_1281\_030\_1301\_030 & 60 \\
segment-15270638100874320175\_2720\_000\_2740\_000 & 60 \\
segment-15349503153813328111\_2160\_000\_2180\_000 & 100 \\
segment-15868625208244306149\_4340\_000\_4360\_000 & 110 \\
segment-16608525782988721413\_100\_000\_120\_000 & 10 \\
segment-17761959194352517553\_5448\_420\_5468\_420 & 0 \\
segment-3425716115468765803\_977\_756\_997\_756 & 0 \\
segment-3988957004231180266\_5566\_500\_5586\_500 & 0 \\
segment-9385013624094020582\_2547\_650\_2567\_650 & 130 \\
segment-8811210064692949185\_3066\_770\_3086\_770 & 30 \\
segment-10275144660749673822\_5755\_561\_5775\_561 & 0 \\
segment-10676267326664322837\_311\_180\_331\_180 & 100 \\
segment-12879640240483815315\_5852\_605\_5872\_605 & 20 \\
segment-13142190313715360621\_3888\_090\_3908\_090 & 0 \\
segment-13196796799137805454\_3036\_940\_3056\_940 & 70 \\
segment-14348136031422182645\_3360\_000\_3380\_000 & 140 \\
segment-15365821471737026848\_1160\_000\_1180\_000 & 0 \\
segment-16470190748368943792\_4369\_490\_4389\_490 & 0 \\
segment-11379226583756500423\_6230\_810\_6250\_810 & 0 \\
segment-13085453465864374565\_2040\_000\_2060\_000 & 110 \\
segment-14004546003548947884\_2331\_861\_2351\_861 & 0 \\
segment-15221704733958986648\_1400\_000\_1420\_000 & 70 \\
segment-16345319168590318167\_1420\_000\_1440\_000 & 0 \\
\bottomrule
\end{tabular}
            \vspace{-3mm}
\caption{\small  Information on the selected and trimmed Waymo Open Dataset~\cite{sun2020scalability}. For each sequence, we select 40 frames starting from the Start Frame.  }
\label{dataset}
    \vspace{-3mm}
\end{table}

\begin{figure}
\centering 
\includegraphics[width=0.47\textwidth]{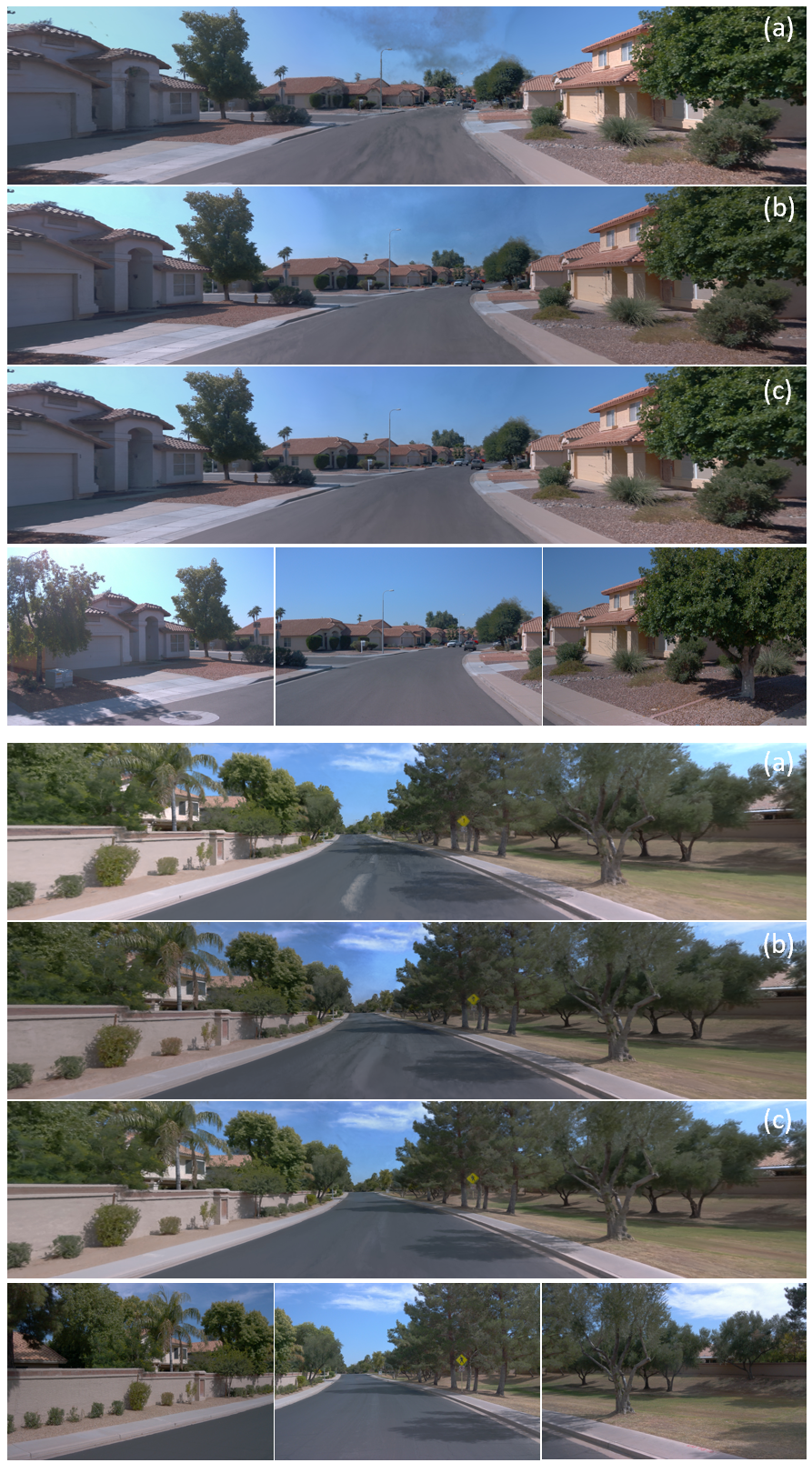} 
\vspace{-3mm}
\caption{\small Qualitative ablation of background rendering. (a) McNeRF w/o pose alignment.(b) McNeRF w/o exposure. (c) Full McNeRF. Last row: target images.}
\label{Fig:ablation} 
\vspace{-3mm}
\end{figure}

\subsection{Multi-Camera Alignment}
This section will introduce the details of our multi-camera alignment algorithm. Let $R_{C_i, t}$ and $T_{C_i, t}$ represents the camera $C_i$'s extrinsic matrix that aligned to vehicle's coordinates at timestamp $t$. $C_0$ is the front camera. The superscript $(V)$ and $(M)$ represents the original vehicle's coordinates in autonomous driving dataset and the coordinates under Metashape's unified space. Then the rotation $R_{C_i, t}$ and translation $T_{C_i, t}$ can be calculated as:
\begin{align*}
R_{C_i, t} &= R^{(V)}_{C_0, 0} (R^{(M)}_{C_0, 0})^{-1} R^{(M)}_{C_i, t} \\
T_{C_i, t} &= \frac{R^{(V)}_{C_0, 0} (R^{(M)}_{C_0, 0})^{-1} (T^{(M)}_{C_i, t} - T^{(M)}_{C_0, 0})}{S} + T^{(V)}_{C_0, 0}, \\
\end{align*}
where $S=\frac{T^{(M)}_{C_0, 1} - T^{(M)}_{C_0, 0}}{T^{(V)}_{C_0, 1} - T^{(V)}_{C_0, 0}}$ is a scaling factor that ensures the aligned space has the same unit length as the real world.

\section{Supplymentary Experiments}

\subsection{Qualitative Ablation Study of Background Rendering}
Figure \ref{Fig:ablation} illustrates the effects of the ablation study on background rendering. It is evident that in the absence of pose adjustment, the rendered results exhibit significant blur and anomalies. Without the intervention of exposure adjustments, there are noticeable changes in brightness at the junctions of different cameras, particularly in the sky. McNeRF, however, successfully avoids these two issues and achieves the optimal rendering outcomes.

\begin{figure*}
\centering 
\includegraphics[width=0.99\textwidth]{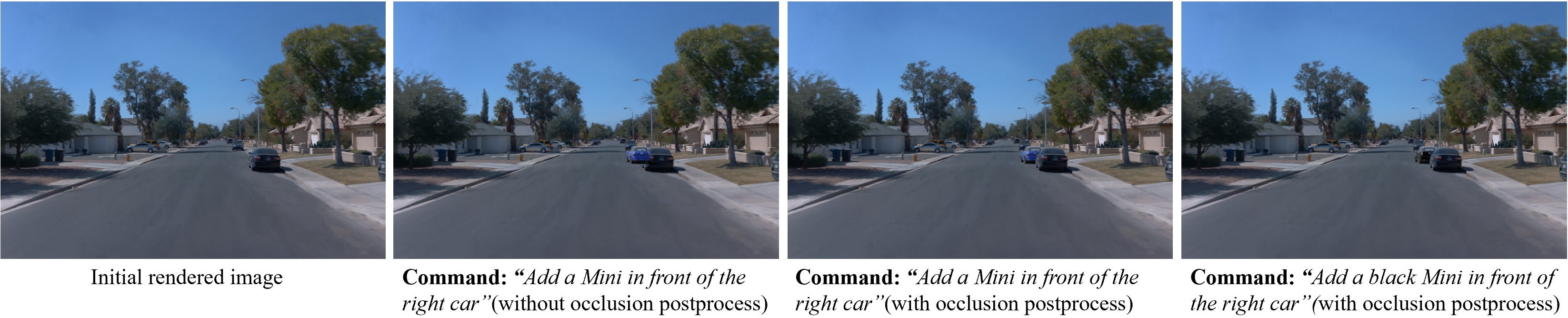} 
\vspace{-3mm}
\caption{\small Qualitative result of occlusion postprocess and the color control for added car.}
\label{Fig:occlusion} 
\end{figure*}
\subsection{Occlusion with Depth Test}
During the process of adding vehicles, there may be instances of occlusion. For occlusions among multiple vehicles to be added, Blender considers this issue during the rendering process. Therefore, we only need to focus on the occlusion between the foreground vehicles and the background objects. The most straightforward method to handle occlusion is determined by the depth map of the foreground and background, respectively. The depth maps of both the foreground and background could be used to choose for each pixel with the lesser depth to be displayed in the front, while the one with greater depth is occluded. However, accurately estimating the background's depth map directly is challenging. The point cloud data in autonomous driving datasets is too sparse, and the depth maps obtained through depth completion are also sparse and excessively noisy, making them unsuitable for pixel-level accuracy in practical use. Here, we combine the sparse depth data from point clouds with the object segmentation method SAM\cite{kirillov2023segment}. SAM can achieve pixel-level accuracy in segmentation results at the image level, without extra finetuning. We first use SAM to obtain different patches in the background image, then identify patches that overlap with the foreground objects. Using the sparse depth map derived from the point clouds, we calculate the average sparse depth within these patches as the depth of each patch. Since the segmentation results of patches often represent a complete instance, and occlusion occurs between instances, it is reasonable to calculate the depth for the entire instance represented by a patch. Subsequently, we create the background's depth map from the depths of these patches and perform occlusion calculations with the depth map rendered for the foreground, presenting each pixel with the lesser depth to finalize the occlusion computation. The results of the occlusion calculation, as shown in Fig. \ref{Fig:occlusion}, illustrate that the added vehicles are occluded by those with shallower depths. This figure also displays the adjustment of the added vehicles' colors.

\begin{figure*}
\centering 
\includegraphics[width=0.99\textwidth]{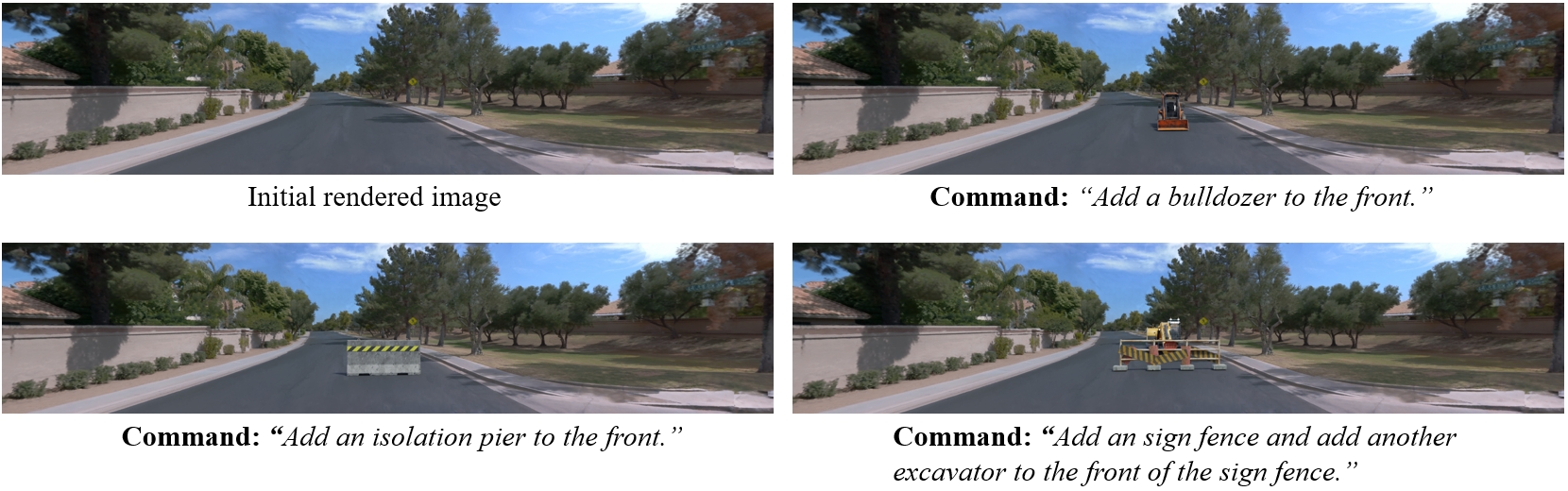} 
\vspace{-3mm}
\caption{\small Qualitative results of rare cases simulation.}
\label{Fig:obstacles} 

\end{figure*}

\subsection{Rare Cases Simulation}
Leveraging diverse external digital assets, ChatSim can simulate rare and challenging-to-collect real-world scenarios within reconstructed existing scenes. Figure \ref{Fig:obstacles} demonstrates ChatSim's ability to emulate rare cases by placing uncommon elements like bulldozers, isolation piers, fences, excavators, and other infrequently encountered situations in reconstructed scenes. This capability enables ChatSim to create rare digital twins for existing collected data, thus fulfilling the need for these specific scenarios.

\begin{table}[]
\centering
\begin{tabular}{c|ccc}
\toprule
Simulation data & AP30            & AP50            & AP70            \\ \hline
            0               & 0.1263          & 0.0366          & 0.0034       \\
            600             & 0.1910          & 0.0878          & 0.0153       \\
            1000            & \textbf{0.2074} & \textbf{0.0930} & \textbf{0.0189} \\
            2200            & 0.2064          & 0.0900          & 0.0182      \\ 
            \hline
\end{tabular}
            \vspace{-3mm}
\caption{Comparison of detection model’s performance with different number of data simulated by ChatSim}
\label{tab:augmentation_2}
\vspace{-3mm}
\end{table}

\subsection{Supplementary 3D Detection Augmentation Experiment}

We conducted 3D detection augmentation experiments under a new setting: we fixed the real data amount from the original dataset at 4200 frames and augmented it with varying quantities of simulation data generated by ChatSim. We continued to use Lift-Splat~\cite{philion2020lift} as the detection model, with results shown in Table \ref{tab:augmentation_2}. It is observed that the use of simulation data significantly enhances the performance of the 3D detection task. As the amount of simulation data increases, the final performance tends to stabilize after a certain point of improvement.


\end{document}